# *eegFloss*: A Python package for refining sleep EEG recordings using machine learning models

Niloy Sikder[a,b], Paul Zerr[a], Mahdad Jafarzadeh Esfahani[a], Martin Dresler[a], Matthias Krauledat[b,*]

[a]Donders Institute for Brain, Cognition and Behaviour, Radboud University Medical Center, Nijmegen, The Netherlands;
[b]Faculty of Technology and Bionics, Rhine-Waal University of Applied Sciences, Kleve, Germany.

[*]**Correspondence:** Matthias Krauledat (matthias.krauledat@hochschule-rhein-waal.de)

**Keywords:** EEG artifact detection, data cleaning, improve sleep autoscoring, TIB detection, Zmax headband, sleep EEG, Python package

## Abstract

Electroencephalography (EEG) allows monitoring of brain activity, providing insights into the functional dynamics of various brain regions and their roles in cognitive processes. EEG is a cornerstone in sleep research, serving as the primary modality of polysomnography—the gold standard in the field. However, EEG signals are prone to artifacts caused by both internal (device-specific) factors and external (environmental) interferences. As sleep studies are becoming larger, most rely on automatic sleep staging, a process highly susceptible to artifacts, leading to erroneous sleep scores. This paper addresses this challenge by introducing *eegFloss*, an open-source Python package to utilize *eegUsability*, a novel machine learning (ML) model designed to detect segments with artifacts in sleep EEG recordings. eegUsability has been trained and evaluated on manually artifact-labeled EEG data collected from 15 participants over 127 nights using the Zmax headband. It demonstrates solid overall classification performance (F1-score ≈ 0.85, Cohen's kappa = 0.78), achieving a high recall rate of ≈94% in identifying channel-wise usable EEG data, and extends beyond Zmax. Additionally, eegFloss offers features such as automatic time-in-bed detection using another ML model named *eegMobility*, filtering out certain artifacts, and generating hypnograms and sleep statistics. By addressing a fundamental challenge faced by most sleep studies, eegFloss can enhance the precision and rigor of their analysis as well as the accuracy and reliability of their outcomes.

# 1. Introduction

Electroencephalography (EEG) is a non-invasive neuroimaging technique that measures the electrical activity of the brain with electrodes placed on various parts of the head. The resultant time-domain signals provide valuable insights into the neural dynamics of various brain regions and cognitive processes in real time. EEG has a rich history of development involving prominent figures from multiple disciplines and spanning over a century. In 1924, German neuropsychiatrist Hans Berger used it to measure human brain activity for the first time [1]. However, his findings were initially met with skepticism, and clinical EEG, as we know it today, took pioneering works from numerous neurophysiologists over the following decades across Europe and the USA to form [2]. Since its inception, EEG has profoundly influenced the field of neuroscience, enabling researchers to identify and study brain rhythms, diagnose neurological disorders, and investigate the neural correlates of cognition and behavior. It has been the driving force behind many key advancements, including the identification of various brain wave patterns, the characterization of event-related potentials (ERPs), and the mapping of sleep stages [3,4]. The invention of EEG, along with its continuous improvement in accuracy and reliability, has paved the way for quantitative research in neuroscience by providing an objective measure to study brain activity. Today, EEG is a foundational tool in neurophysiology, with its applications extending beyond the realms of neuroscience into psychology, computer science, engineering, brain-computer interface (BCI) research, robotics, and even business, sports, and art [5–11].

Neuroscience is no longer confined to EEG in its quest to study the brain and unravel the workings behind its diverse functions. Compared to EEG, modern imaging techniques such as functional ultrasound (fUS), magnetoencephalography (MEG), positron emission tomography (PET), and functional magnetic resonance imaging (fMRI) offer higher spatial resolution (in the case of MEG, comparable spatial resolution) along with more precise source localization, deep-brain imaging capabilities, and whole-brain coverage [12–15]. However, each of these techniques comes with its own set of limitations. The primary advantage of EEG lies in its ability to measure the brain's electrical activity directly with millisecond-level precision [16]. Modern EEG systems are also highly customizable, compact (often portable), cost-effective, silent in operation, less environment-dependent, and capable of recording high-density EEG (hdEEG) signals. As a result, despite the prevalence of advanced imaging technologies, EEG firmly holds its place in neuroscience and its subdomains, particularly in sleep research [17]. That said, EEG has some inherent limitations, primarily its susceptibility to various artifacts. An EEG artifact is any measured signal that does not represent brain activity. EEG signals are often contaminated by eye movements, cardiac activity, movements of seemingly unrelated or independent muscle groups, poor grounding, electromagnetic interference, power-line noise, and environmental factors [18]. While some artifacts can be mitigated with precise filtering techniques, others completely obscure brain waves and pose a serious challenge during analysis, a problem that remains prevalent without a universal or general solution [19].

Choosing a particular technology always involves compromises regarding the size and shape of the observed area, temporal and spatial resolutions, and other physical constraints, such as cost, setup time, and required expertise [20]. To counter this, studies often combine multiple techniques so that the technical limitations of one can be compensated for by the other. EEG is often paired with fUS, MEG, PET, and fMRI to study seizure, epilepsy, cognition, metabolic and hemodynamic responses, and other phenomena [21–24]. The resultant multimodal data provides a wealth of information on the phenomenon in question and contains valuable insights. A similar approach is evident in sleep research, where EEG is paired with electromyogram (EMG), electrooculogram (EOG), and electrocardiogram (ECG) signals to form a coalition called polysomnography (PSG) to record sleep. Although EMG, EOG, and ECG are not considered neuroimaging techniques, they provide information on muscle activity, eye movements, and cardiac rhythms (respectively) during sleep, which, together with concurrent brain waves, offer a comprehensive means to study sleep architecture. PSG, therefore, has become a standard measurement protocol in sleep research, used to identify the fundamental components of sleep, including sleep stages, slow waves, spindles, and anomalies.

The process of identifying sleep stages—wake (W), rapid eye movement (REM), and non-rapid eye movement (NREM) stages 1–3 (N1, N2, and N3, respectively)—through visual inspection of PSG data is known as *manual sleep scoring*. The N3 sleep stage is also known as *slow-wave sleep* (SWS) or *deep sleep*. Manual sleep scoring is typically performed by an expert sleep scorer, who examines 30-second (or 1-minute) segments (referred to as an epoch) of parallel EEG, EOG, and EMG or ECG data and assigns the appropriate (or most likely) sleep stage based on the standard American Academy of Sleep Medicine (AASM) guidelines [25,26]. After reviewing the entire night (or sleep period), this process usually results in two sets of labels: epoch-wise sleep stages (or *sleep scores*) and arousals (brief interruptions in sleep that may not lead to full wakefulness or awareness). The first set of labels is often visualized as a step chart against time, illustrating the progression of sleep stages throughout the night and highlighting their transitions. This graphical representation, known as a *hypnogram*, is a widely used tool for analyzing sleep architecture.

EEG is the most vital modality of PSG for sleep scoring, as it contains the majority of the distinct features associated with individual sleep stages. In contrast, EMG, EOG, and ECG provide supplementary (yet essential) information that assists scorers in more accurately identifying REM sleep and arousals [27]. Consequently, most PSG systems are typically equipped with a higher number of EEG channels (ranging from 12 to 64) compared to just 2–3 channels for each of the other modalities. This redundancy makes reliable manual sleep scoring possible, even if some EEG channels contain irremovable artifacts. Hence, manually identified sleep scores are widely regarded as highly accurate, as they involve expert inspection of the data "by hand," making it the gold standard in most sleep studies. Moreover, the impact of EEG artifacts is largely minimized through manual review and the availability of redundant EEG channels. In cases where all EEG channels are affected by artifacts, scorers typically label those epochs as such, allowing them to be excluded from subsequent analysis and further reducing the influence of artifacts.

Nevertheless, for all its advantages, manual sleep scoring is a labor-intensive and time-consuming process. According to two experienced sleep scorers at the Donders Sleep & Memory Lab, scoring an 8-hour night takes between 30 and 60 minutes. However, other studies have reported scoring times of up to three hours per recording [28]. Moreover, being entirely reliant on human judgments, it is prone to inter-scorer variability and subjective biases [29]. Furthermore, recording PSG is costly and requires special laboratory environments (although recording at home is possible [30]). Despite these challenges, PSG-based manual sleep scoring remains the most reliable method for sleep staging and is routinely performed in major PSG-based sleep studies.

Recent advancements in Machine Learning (ML)—driven by exponential increases in computing power, storage capacity, and the development of sophisticated algorithms—have enabled machines to tackle and solve numerous complex problems. Over the past two decades, ML models have been applied across nearly every scientific domain that involves data-driven decision-making, with their influence continuing to expand. Neuroscience, a field heavily reliant on neurological data, is no exception [31]. In this domain, ML algorithms have been employed to identify patterns of brain activity associated with specific cognitive states, emotions, and tasks [28], further our understanding of brain connectivity [32], and decode neural signals [33], to name a few. However, they are most frequently used to develop computer-aided diagnosis (CAD) methods. In neuroscience, CAD applications include the detection, diagnosis, and monitoring of neurological diseases (such as epilepsy, multiple sclerosis, Alzheimer's, and Parkinson's) [34] and mental health conditions (such as anxiety, depression, and schizophrenia) [35].

ML models have been widely adopted for automatic sleep-stage identification in sleep research involving healthy humans [36]. This process is similar to manual sleep scoring, except that an ML algorithm replaces the human scorer. In this article, we will use the term *autoscorer* to refer to an ML model capable of automatically distinguishing between sleep stages (based on data from PSG, EEG, or other modalities), *autoscoring* to the act of applying an autoscorer, and *autoscores* to denote the resultant set of sleep scores. Typically, autoscorers are supervised learning (SL)-based models trained on existing manual sleep scores. Autoscoring sleep data promptly solves the first two drawbacks of manual sleep scoring—extensive time and labor requirements—as it can score an entire night in mere seconds (or, at most, 1–2 minutes) and does not require human input. As sleep studies grow in scale, many researchers are turning to autoscorers such as *U-Sleep* [37] and *YASA* [38] to score PSG data. Although not perfect, these models demonstrate substantial agreement with manual scores [39,40]. However, autoscorers usually do not explicitly check for specific artifacts in EEG signals, which can affect their outputs. Nonetheless, they can still leverage the redundancy provided by multiple EEG channels to partially mitigate the impact of artifacts on the resulting autoscores. For these reasons, PSG-based autoscorers are gradually replacing manual scorers in sleep research, even though the implications of this shift are still unclear [41]. That said, even with high-quality EEG data, the accuracy of existing autoscorers still falls short of human scorers, particularly in identifying the N1 sleep stage. Consequently, the development of new and enhanced autoscorers remains an active area of research.

Turning our attention to the last two drawbacks of PSG—cost and lab dependency—modern wearable devices (or *wearables*) with EEG recording capabilities may offer a solution to them [42]. Notably, wearable EEG headbands equipped with 2–8 EEG channels, photoplethysmography (PPG) sensors, thermometers, pulse oximeters (SpO$_2$), and accelerometers (ACC) have become increasingly common in recent years [43]. Examples of such wearables include *Dreem* [44], *Muse S* [45], *Zmax* [46], *SleepLoop* [47], *SmartSleep* [48], *Sleep Profiler* [49], *Bitbrain Ikon* [50], and *OpenBCI* [51]. While most of them are aimed at consumer use, some—such as Muse S, Zmax, SleepLoop, and Bitbrain Ikon—have been used in sleep studies, and their outputs have been validated against traditional PSG recordings [52–56]. The primary advantage of using wearable EEG headbands in sleep research is their convenience. Compared with PSG, they are affordable, self-applicable, easy to use, relatively robust, and well-suited for frequent and long-term sleep monitoring with minimal maintenance. However, as previously noted, there are trade-offs—the primary being the limited number of EEG channels and the secondary being the absence of standardized guidelines, particularly for sleep scoring. Most headbands collect signals from the forehead (and, in some cases, the back of the head), resulting in narrower coverage compared to PSG. Additionally, most headbands use dry electrodes (Zmax and SleepLoop are notable exceptions), which typically provide lower signal quality compared to the wet electrodes used in PSG systems [57]. Furthermore, headbands typically do not record EMG, EOG, or ECG, which are integral parts of PSG, making their scope of surveillance even narrower. That said, wearable EEG headbands introduce alternative modalities—such as PPG, SpO$_2$, ACC, and body or skin temperature—usually absent in PSG systems. Whether these alternate modalities can be leveraged for clinical sleep research remains to be seen [58].

Sleep scoring of headband data is typically carried out using device-specific autoscorers. Similar to PSG-based autoscorers, they are usually also trained on manual sleep scores derived from (parallel) PSG recordings. These autoscorers vary in terms of their complexity, base model, performance, number of target classes (ranging from two to six), and, most importantly, input data modalities [59]. As previously discussed, headbands usually offer EEG, PPG, ACC, SpO$_2$, and temperature readings during sleep. While various combinations of input modalities are possible (and have been used), EEG and ACC-based autoscorers generally yield better performance than the others [60]. However, PPG, SpO$_2$, and temperature are also being tested for autoscoring [61]. Nevertheless, EEG signals remain central for wearables-based autoscorers as well. Since they have fewer EEG channels, the performance of these autoscorers largely depends on those few channels having good-quality data, which, unfortunately, is often not the case. EEG data from wearables is particularly susceptible to noise and artifacts, primarily due to the use of dry electrodes, which are often less securely affixed to the contact points without any conductive gel to improve conductivity, further degrading signal quality [57]. Therefore, these autoscorers often become overly reliant on a few EEG channels that are prone to contamination by artifacts. Moreover, unlike manual scoring, autoscoring does not involve human inspection to filter out artifacts. Consequently, these artifacts typically persist in the EEG data and, posing a significant challenge to the autoscorers, actively hinder their classification performance [62,63].

In this article, we introduce a novel ML model called *eegUsability* that can automatically identify sleep EEG segments contaminated with various types of artifacts, allowing for either the removal of their corresponding sleep scores (or the exclusion of these segments from being autoscored altogether). eegUsability is a Light Gradient-Boosting Machine (LightGBM)-based model and was trained and validated on 127 nights of manually artifact-labeled EEG data collected from 15 participants using the Zmax headband [64]. Although eegUsability was developed using Zmax EEG data, it can be used to check sleep EEG data from other devices. To bring it out of the pages of this article and into practical use, we have developed a Python package named *eegFloss*, which provides a fast and straightforward way to apply the *eegUsability* model to new data. Although artifact detection is the primary goal of eegFloss, it includes additional features. Notably, it incorporates *eegMobility*, a model designed to identify non-sleep periods in EEG recordings—such as the time before going to bed and after waking up—based on the degree of mobility, which can be used for automatic time-in-bed (TIB) detection. To the best of our knowledge, eegUsability and eegMobility are the first models of their kind. We found no prior ML models in the literature trained on 100+ nights of manually artifact-labeled sleep EEG data from multiple (healthy human) participants for automatic artifact detection, and no existing models have been specifically trained on Zmax ACC data for automatic TIB detection.

eegFloss can also aggregate the outputs of eegUsability (or *usability scores*) with existing sleep scores to generate *artifact-rejected scores*, filter a type of (high-frequency, low-amplitude) artifact, create visualizations, and calculate sleep statistics. We also describe different variations of the models, how and when to use them, and how these models (and the package) can be adapted to meet the requirements of different projects. The eegFloss package, its base models, and other accompanying scripts are open-source, free to use, and can be modified as necessary (see the Code availability section). It has already been used on more than seven sleep EEG datasets without any major issues. These (visual) outputs have been presented as part of this article (see the Supplementary materials section). The presence of artifacts in EEG data is inevitable, and if not dealt with properly (i.e., identified and then filtered or rejected), their adverse effects can misdirect analyses and sway study outcomes. Using the eegFloss package to process them before the main analysis will undoubtedly make the results of any sleep study more flawless and cogent.

## 2. Recent artifact detection methods for sleep EEG

Since the advent of EEG, artifacts have been part of it, as have the relentless efforts of researchers to mitigate them. Early methods for detecting EEG artifacts relied heavily on various filtering techniques based on digital signal processing (DSP) algorithms, which gradually became more effective as DSP technology advanced. A comprehensive 2018 review article cataloged these methods, tracing their evolution as far back as the 1970s [65]. In the last two decades, artifact detection algorithms have generally fallen into a few broad categories based on their underlying methodologies: regression, wavelet transform, empirical mode decomposition (EMD), adaptive and non-adaptive filtering, and blind source separation (BSS) techniques. The BSS category includes methods such as principal component analysis (PCA), independent component analysis (ICA), canonical correlation analysis (CCA), and sparse component analysis (SCA) [19]. More recent studies have often combined these approaches to create hybrid methods, which tend to be more effective than individual techniques. Nevertheless, automatic artifact detection methods for sleep EEG data can be broadly classified into two groups: statistical thresholding-based methods (discussed above) and ML-based methods. Although this article falls into the latter group, it is essential to acknowledge the contributions of statistical approaches. This section highlights some of the most notable studies from the past decade attempting to tackle this longstanding challenge.

A 2016 study developed an automatic artifact and arousal detection method using power spectral density (PSD) analysis and adaptive thresholds [66]. The method was evaluated on 60 full-night sleep recordings from four datasets, manually scored by six human raters. The results showed moderate to substantial agreement between the method and human ratings, with Cohen's Kappa ($\kappa$) values ranging from 0.50 to 0.72. Another 2016 study introduced an automatic artifact removal method focusing on EOG and EMG artifacts in single-channel sleep EEG data [67]. It used cross-correlation thresholding for artifact detection and normalized least-mean squares (NLMS) adaptive filtering for artifact removal. The authors reported a 14% improvement in sleep autoscoring accuracy after applying their method.

A 2019 study proposed an automatic artifact detection method for single-channel sleep EEG recordings [68]. The researchers implemented 14 different detection algorithms, optimized them using 32 recordings, and validated their method on 21 healthy subjects and 10 patients from multiple laboratories. The best-performing approaches relied on fixed thresholds for EEG slope, high-frequency power (25–90 Hz or 45–90 Hz), and residuals from adaptive autoregressive models, achieving ≈90% sensitivity and specificity. A 2023 study introduced a MATLAB-based, semi-automatic artifact removal routine tailored for hdEEG recordings [69]. The method features a graphical user interface (GUI) that guides users through eight iterations of artifact detection using four signal quality markers, enabling manual inspection and removal of outliers across all channels. However, its main limitations are the need for expert users to guide the process (hence, not entirely automatic) and its focus on hdEEG (64+ channels), which may lead to suboptimal results when applied to data with fewer channels.

A 2024 study introduced a customizable automated cleaning tool for multichannel sleep EEG recordings, addressing the challenges of manual artifact detection in hdEEG [70]. The method included channel-wise artifact detection, segment-wise repair via interpolation, and visualization tools for monitoring data quality. The system was evaluated on 560 overnight recordings from 371 individuals and achieved high accuracy (92–95%) but low precision (39–48%), indicating the over-detection of artifacts. Another 2024 study applied multitaper spectral analysis to compute power correlations across six EEG leads for automatic artifact detection [71]. The method identified low-correlation segments as artifacts using a local moving window and global thresholding approach. Evaluated on 9,641 PSG recordings, the system achieved 80% recall compared to expert-annotated ground truth. The main limitation of these methods is their reliance on fixed thresholds, which reduces their generalizability across different participants, situations, and datasets [72].

Lastly, a 2025 study compared manual and automatic artifact detection in 252 healthy volunteers' sleep EEG recordings, focusing on PSD estimates [73]. The method used Hjorth parameters for automatic artifact detection and found that extreme artifacts caused the most distortions. While manual and automatic detections had only moderate agreement, the overall PSD estimates remained highly similar across methods, suggesting that visual artifact detection may not be necessary for large-scale sleep EEG databases.

In contrast, ML-based approaches adapt to variations in data, making them more robust and widely applicable. However, ML methods for EEG artifact detection are relatively rare, primarily due to the scarcity of artifact-labeled EEG datasets. An early attempt was made in 1989, where researchers applied clustering and SL to detect muscular artifacts in EEG signals [74]. The method used Autoregressive (AR) modeling with Burg and Levinson-Durbin algorithms, followed by K-means clustering. The resulting clusters were then used to train three classifiers—Fisher's Linear Discriminant (FLD), Q-NN rule, and Mahalanobis Distance. Among them, the last two achieved high concordance (≈92%) but had variable degrees of sensitivity. Additionally, ocular artifacts were detected by correlating EEG and EOG signals. ML advancements in the last two decades have led to the emergence of deep learning (DL) and ensemble-based models, surpassing earlier approaches concerning their robustness and classification accuracy.

Two studies from 2018 used advanced ML models to identify artifacts in sleep EEG signals. The first introduced a DL-based automated classification system for sleep autoscoring and EEG artifact detection in mouse data [75]. The model used convolutional neural networks (CNNs) to classify EEG and EMG signals into W, NREM, and REM sleep stages and detect the presence or absence of artifacts in 4-second epochs. Evaluated on manually sleep-scored and artifact-labeled 82 recordings from 22 mice, the model achieved an accuracy of 96% for sleep stage classification and 85% for artifact detection on the validation set. The other study proposed an automatic artifact detection method for multichannel sleep EEG using a Random Forest classifier [76]. It involved adaptive segmentation, feature extraction, and classification to identify artifacts. Evaluated on EEG data of short durations (10–20 minutes) from 14 subjects, the approach showed a 16% improvement in the F1-score for wake EEG, with increases of 9%, 5%, and 16% for REM, N2, and N3, respectively, compared to baseline methods. The study noted that rare or atypical artifacts were more difficult to classify.

As perceived from this discussion, the existing ML methods for sleep EEG artifact detection are neither focused on overnight data from healthy humans nor tested on diverse datasets or a large number of recordings. This sizable research gap is the primary motivation behind our undertaking of the challenge and development of the eegFloss package.

## 3. Core models

The eegFloss package is built around two core ML models: eegUsability and eegMobility. eegUsability is trained to identify artifact-contaminated EEG segments, while eegMobility assesses the degree of a participant's mobility or movements throughout the night. In this section, we provide detailed descriptions of the formulation and performance of these models.

### 3.1. Model eegUsability

eegUsability is a LightGBM-based SL model [77]. Given a single-channel EEG along with its corresponding tri-axial accelerometer (ACC) signals, the model segments the data into 10-second epochs and determines whether each epoch is "good" or contains artifacts. In this article, the terms *good*, *usable*, and *scorable data* have been used interchangeably to refer to sleep EEG data that are (visually) free from artifacts and can be used for autoscoring. Since our primary focus is to preserve as much usable data as possible, we aim to emphasize this class and minimize its misclassification rate as much as possible. We will also interchange the terms "noise" and "artifact" in the rest of the article, as some artifact classes do not show any discernible patterns and their origins are still unclear. As with most SL models, the development of eegUsability included several key steps, namely, data preparation, class balancing, feature extraction, model training, and performance evaluation. The entire training and evaluation workflow for the eegUsability model is illustrated in Figure 1.

#### 3.1.1. eegUsability: training data preparation

To prepare the training data for the eegUsability model, we utilized the Donders 2022 dataset, which includes overnight EEG recordings from healthy young participants (mean age: 21.5 ± 3.78 years) using the Zmax headband [53]. The Lite version of the Zmax headband provides two-channel forehead EEG (denoted as *EEG Left* and *EEG Right*) using wet (gel-type) patch electrodes, PPG, and tri-axial ACC readouts at a sampling rate of 256 Hz [78]. For this study, we visually inspected 181 nights of EEG and ACC data from 15 participants and annotated the presence of artifacts. The annotation was performed based on the visual characteristics of the EEG signals after bandpass filtering (0.3–30 Hz). Below, we detail the various artifacts encountered in the Zmax EEG data.

#### 3.1.1.1. Artifact labeling

The artifact labeling process we employed was akin to manual sleep scoring, except that we identified specific artifact types instead of sleep stages, noting the precise start and end times (in seconds) of each artifact rather than assigning scores for each 30-second (or 1-minute) epoch. We identified artifacts in each EEG channel (separately) by inspecting the corresponding channel's windowed spectrogram (a visual representation of the distribution of signal power across frequencies and over time [79]), the tri-axial ACC data, and the raw time-domain signal. We identified four primary (or most recurring) artifacts in Zmax EEG data. Together with usable data, these constitute five distinct classes:

- *Good Data* (*Class 0*): typical (or expected) sleep EEG data with amplitudes between −100 µV and 100 µV, free from any discernible artificial or sleep-unrelated patterns. These signals may contain universally recognized sleep-related features, such as sleep spindles, K-complexes, and alpha and delta waves, which are characteristic of different sleep stages. Also, known EEG artifacts that fall within this range were not distinguished, as sleep scoring can usually be performed despite their presence.
- *No Data* (*Class 1*): an EEG signal that remains unchanged at a constant amplitude (which may not necessarily be 0 µV), showing no meaningful fluctuations over time. This typically results from a disconnected (or loosely connected) EEG electrode or its reference, leading to a flat, non-informative signal. These segments exhibit no discernible power in the corresponding spectrogram, implying the absence of neural activity.
- *High Noise* (*Class 2*): an EEG signal with amplitudes far exceeding the normal range, indicating contamination by external sources rather than representing brain activity during sleep. However, such high-amplitude signals can also be associated with wakefulness or arousal-related movements. ACC signals were used to differentiate noise from movement artifacts—if any movement was detected in the corresponding ACC output, the segment was not classified as noise. These segments exhibit high power across a broad frequency range in the spectrogram.
- *Spiky Noise* (*Class 3*): a type of high-frequency, low-amplitude artifact that obscures underlying brain activity. While its amplitude may vary, this noise is characterized by distinct spectral features, appearing as clearly discernible high power at ≈8 Hz and its harmonics in the corresponding spectrogram. This artifact can interfere with EEG analysis by distorting low-amplitude sleep-related oscillations.
- *M-shaped Noise* (*Class 4*): a distinctive periodic artifact characterized by a recurring M-shaped waveform, with amplitudes ranging between around −100 µV and 100 µV and a period of 4–5 seconds (frequency: 0.2–0.25 Hz). The M-shape becomes more pronounced when the signal is filtered between 0.3 Hz and 30 Hz, as shown in Appendix 1, Figure 1.1(a). Due to its amplitude being within the range of typical sleep EEG signals, it can be challenging to distinguish from usable EEG data in spectrograms. This artifact exhibits visual similarities to several instances of sweat and respiratory artifacts reported in the sleep EEG literature [80–82]. However, it might also be a device-specific anomaly, co-occurring in both EEG channels and in phase, ruling out rolling eye movements as a potential cause.

Representative samples of these classes are depicted in Figure 2, alongside the windowed spectrogram of a single-channel overnight sleep recording using Zmax, with specific segments marked where they occurred.

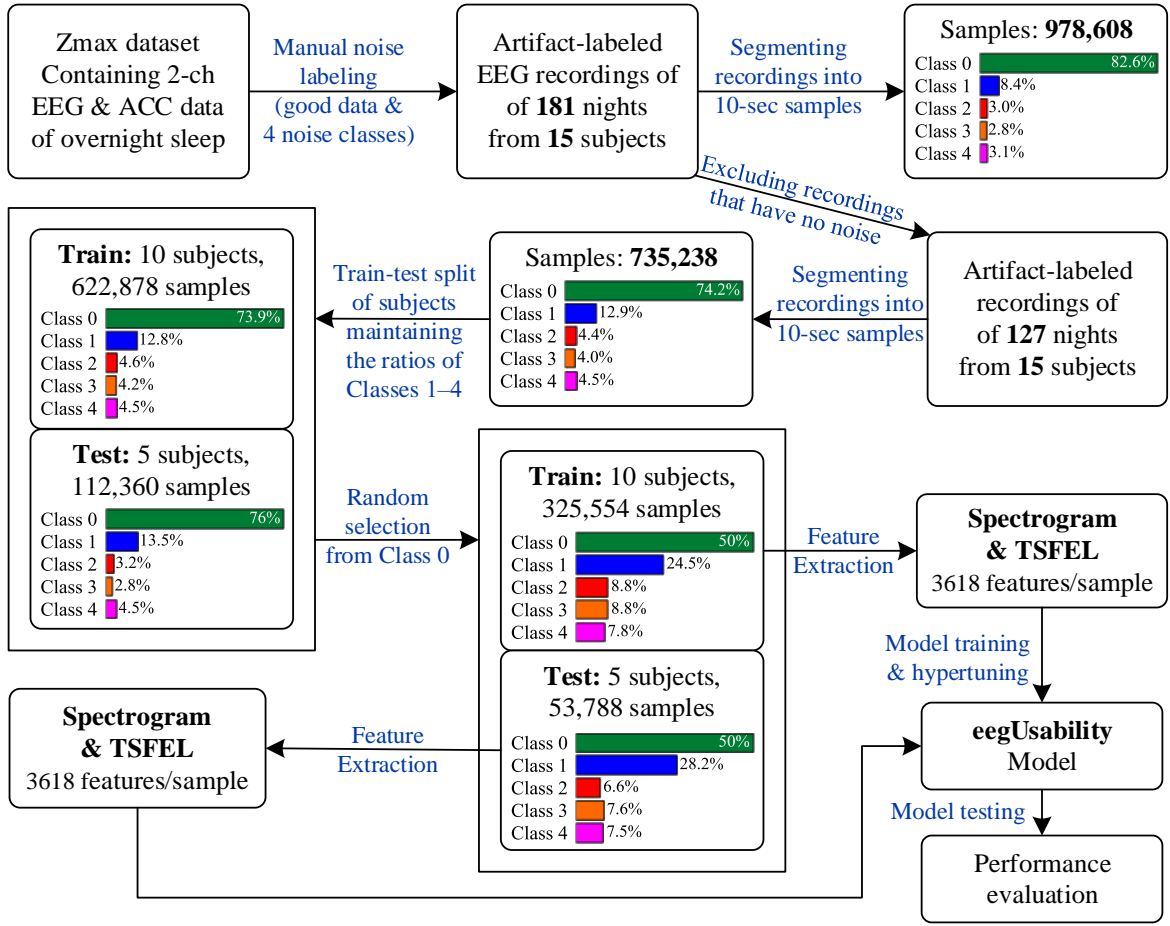

**Figure 1:** Detailed workflow of training and evaluating the eegUsability model, including sample preparation, class ratio balancing, and feature extraction.

### 3.1.1.2. Labeling compound artifacts

In addition to the primary artifacts described, we identified many less frequent ones as well as compound artifacts—where multiple artifacts occurred simultaneously within a single segment. Spiky Noise was often found superimposed on other noises. To address this issue, we implemented secondary labeling—when a compound artifact was encountered, we labeled the most disruptive one as the primary artifact and the other as a secondary artifact. A few examples of such cases have been presented in Appendix 1, Figure 1.1(d–f). Most notably, *Class 5* and *Class 6* are variants of the M-shaped Noise with higher amplitude ranges (±400μV and ±800μV, respectively). Spiky Noise can be found superimposed on other classes—No Data, High Noise, and M-shaped Noise—marked as *Class 13*, *Class 23*, and *Class 43*, respectively.

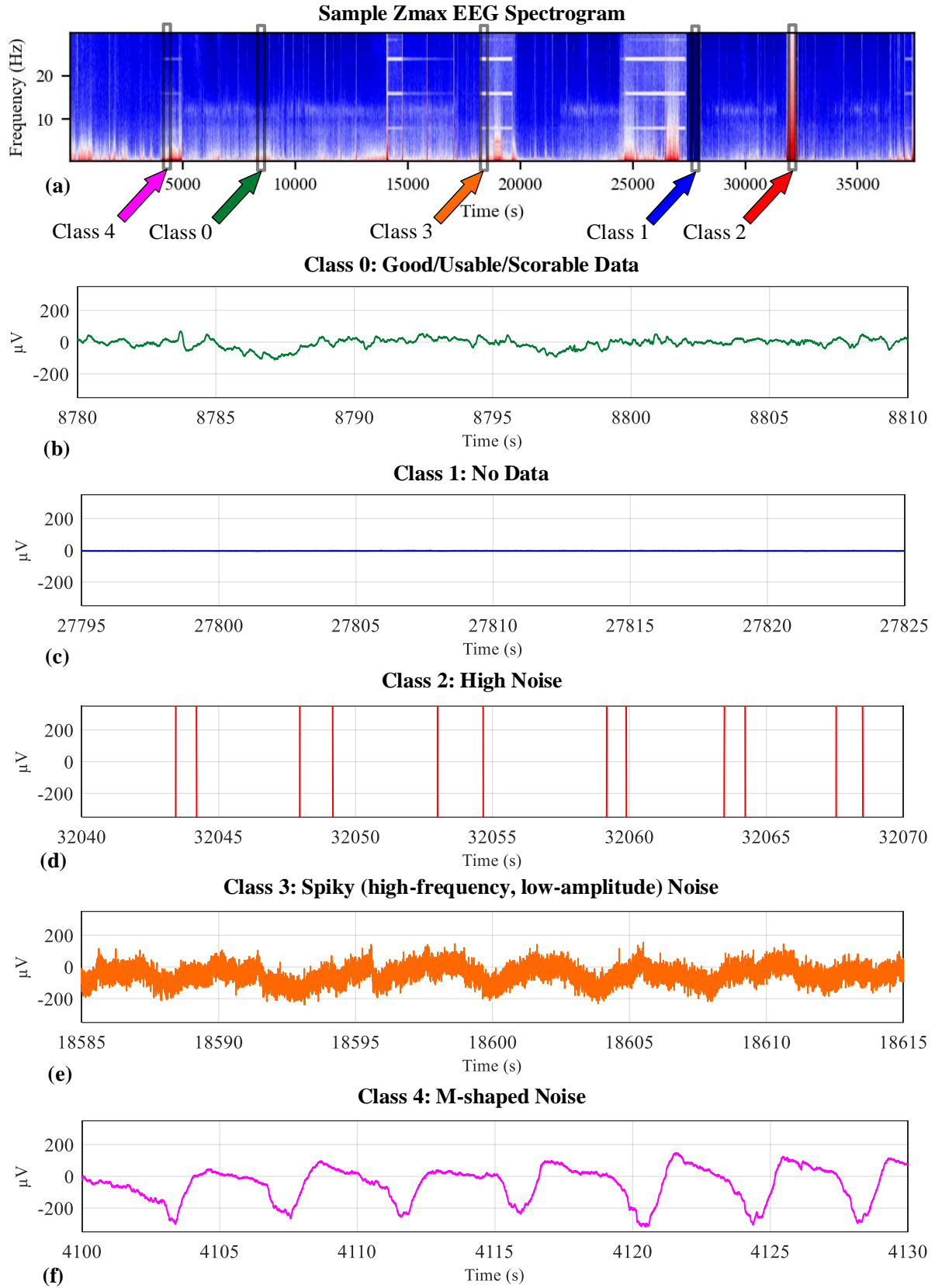

**Figure 2: (a)** A windowed spectrogram (blue: low power, red: high power) of a sample Zmax EEG channel, highlighting segments containing different artifacts. The corresponding time-domain representations of these segments are shown for **(b)** Good Data, **(c)** No Data, **(d)** High Noise, **(e)** Spiky Noise, and **(f)** M-shaped Noise.

### 3.1.1.3. Combining artifact labels

Apart from the artifacts described in Sections 3.1.1.1 and 3.1.1.2, several additional artifact types were present in the Donders 2022 dataset; however, these were even less frequent. Since this study primarily focused on identifying usable data rather than more specific artifact types, we simplified the labeling process to ensure a robust classifier with sufficient data in each class. To achieve this, we merged several artifact labels based on their characteristics:

- Class 5 and Class 6 were merged with High Noise since they exhibited similar properties.
- Class 13, Class 23, and Class 43 were merged with Spiky Noise, High Noise, and M-shaped Noise, respectively, as their features aligned with the definitions of these primary noise categories.
- Other artifact labels were either merged accordingly or deemed too negligible to be included and were therefore ignored.

After combining the labels, we obtained a final dataset containing raw EEG and ACC data, along with corresponding labels indicating whether the segments were usable for autoscoring or contained one of the four primary artifacts.

### 3.1.1.4. Sample preparation

After determining the class labels, we prepared individual samples from the Zmax recordings to train the model. In this step, we applied a sliding window to segment the raw data to generate samples. As previously mentioned, we prioritized minimizing the loss of good data and, thus, initially considered using five-second windows. However, given that the M-shaped Noise typically exhibits a period of around 4 seconds (with a fundamental frequency of ≈0.25 Hz), there was a risk that such a short window might not encompass an entire cycle of this artifact, potentially missing its distinctive characteristics. Consequently, we opted for ten-second windows, sufficient to effectively capture the inherent properties of all classes. Thus, for usability (and mobility) detection, ten seconds of data constitute an epoch, referred to hereafter as a *10s-epoch*, to distinguish it from an epoch for sleep scoring, which is usually 30 seconds. Additionally, eegUsability takes both EEG (single-channel) and parallel (tri-axial) ACC data into account to detect specific artifact classes within each 10s-epoch, and we will refer to this combined 10-second data as a *sample* in further descriptions.

From the previous step, we obtained the exact start and end times of artifacts in the data, marked in seconds. In this step, we established the definitive label for each sample using two key rules. Firstly, a sample is labeled according to the class representing the majority of its data—the sample is assigned to the class that is most prevalent within it. Secondly, in the event of a tie— where two or more classes had equal representation—we applied a predefined order of precedence to resolve the deadlock, prioritizing them as follows (from highest to lowest): Class 0, Class 2, Class 4, Class 3, and Class 1. These rules effectively resolved any ambiguities in label assignment, ensuring each sample was accurately categorized according to its dominant artifact type. The outcomes of this step are a set of samples and their corresponding (single) artifact labels.

### 3.1.2. eegUsability: class balancing

Segmenting the entire labeled dataset yielded approximately one million samples from 181 nights. However, the proportions of class samples for the entire dataset (referred to as the *class ratios*) were significantly imbalanced, as shown in Figure 1, with usable data comprising more than 82% of the dataset, leaving the other classes substantially underrepresented. Such an imbalance can lead to a model biased towards the major class, as the minor classes might not be adequately comprehended by the classifier. To address this issue, we implemented two random undersampling (RUS) techniques, which usually lead to better model performance [83]. First, we excluded 54 of the 181 recordings that contained no noisy data, effectively reducing many samples from Class 0 without impacting other classes. Despite this, the class ratios remained unbalanced. The second undersampling operation was performed after training and test subsets were created. As the names suggest, the first was used to train and hypertune the model, and the latter to test its performance on unseen data.

Although we initially planned to implement a random 80:20 (train-test) split, two critical factors influenced our approach: firstly, ensuring that all classes were adequately (and proportionately) represented in both sets, and secondly, preventing data from the same subject from featuring in both sets to reduce biases—this ensures that the test scores accurately reflect the model's true performance). The second constraint complicated the implementation of the first, as we needed to find a split among the subjects that preserved the original class ratios as closely as possible. Ultimately, by considering the subject-wise class ratios of Classes 1–4 and exploring different combinations, we divided the subjects into groups of 10 and 5 for the training and test subsets, respectively, achieving class ratios that closely match those of the original dataset. The overall ratio of the training and test samples was approximately 86:14.

In our second step of undersampling, we counted the total number of samples from Classes 1–4 and randomly selected the same number of samples from Class 0 of the training subset (and later, the test subset as well). This approach resulted in a more balanced dataset, where 50% of the samples were usable, ≈25% were from No Data, and the remaining 25% were distributed among the other three classes, each represented almost equally. While this would not eliminate biases entirely (as usable data still accounted for half of the dataset), it ensured that all classes were represented sufficiently for the model to learn their distinct properties. Additionally, to minimize the loss of usable data, this class was well-represented, reflecting its diverse characteristics, and a slight bias towards this class was deemed acceptable. Detailed numerical breakdowns after each step of adjustment have been provided in Figure 1.

### 3.1.3. eegUsability: feature extraction

We extracted two types of features from each 10s-epoch: spectrogram features and a set of statistical (along with temporal and spectral) features, which we will refer to as *TSFEL* features after the Time Series Feature Extraction Library, which was used to extract them [84,85]. At this point, each sample comprised four signals—EEG and tri-axial ACC. However, as our interest was primarily in detecting movement, irrespective of its direction, we calculated the magnitude of the tri-axial ACC signals using the Euclidean norm:

$$A_{norm}(t) = \sqrt{(A_x(t))^2 + (A_y(t))^2 + (A_z(t))^2} \qquad (1)$$

where $A_x$, $A_y$, and $A_z$ are the individual accelerations along the X, Y, and Z axes, respectively, at any given data point $t$. This aggregation resulted in each sample containing two main signal segments—EEG and $A_{norm}$. Below, we detail how various sets of features were extracted and utilized to train the model.

### 3.1.3.1. Spectrogram features

A spectrogram is a visual representation of the frequency content of a signal over time. It is commonly used to analyze time-varying frequency information, most commonly, the power at different frequency bands. Spectrograms can be helpful in visually identifying patterns, distinguishing different sources, and detecting anomalies in signals. As mentioned in Section 3.1.1.1 and seen in Figure 2, three of the four primary artifacts are discernible from the signal's windowed spectrogram. Thus, we directly used this type of feature for classification.

Considering $x(t)$ as a time-series signal sampled at a rate of $f_s = 256$ Hz over 10 seconds, resulting in a total of $256 \times 10 = 2560$ data points, the Fast Fourier Transform (FFT) of $x(t)$ can be expressed as:

$$X(t, f) = \sum_{n=1}^{2560} x(n) w(n-t) e^{-j2\pi fn/f_s} \qquad (2)$$

where $x(n)$ represents the signal at discrete time $n$, $w(n-t)$ is a window function centered around time $t$, and $f$ is the frequency. The spectrogram $S_p$ of $x(t)$ is constituted by the magnitude squared of the STFT coefficients, which can be expressed as:

$$S_p(f, t) = |X(t, f)|^2 \qquad (3)$$

For the EEG signals in question, calculating the spectrogram of each signal segment results in a feature map shaped as $(11 \times 129)$. These features were flattened (concatenated row-wise) and combined with the features coming from the ACC of the same shape, resulting in $11 \times 129 \times 2 = 2838$ spectrogram features ($F_{sp}$) for each sample.

### 3.1.3.2. TSFEL features

To enhance the model's ability to discern the underlying structure of the samples, we calculated their statistical properties using the TSFEL library (version 0.1.4), incorporating this data into the model training process. The TSFEL library is designed to extract a comprehensive set of statistical, temporal, and spectral features from signals. For an exhaustive list of the features it provides, please refer to the library's documentation [86]. In our analysis, TSFEL generated 390 distinct features from each signal segment, totaling 780 features per sample ($F_{ts}$), considering both EEG and $A_{norm}$. These features were then concatenated with $F_{sp}$, culminating in 3618 features per sample. This extensive feature set aimed to provide a robust foundation for the model to learn from and improve its predictions.

In addition to the spectrogram and TSFEL features, we extracted other information from the samples, including frequency-domain data, power properties via methods such as Welch and Burg, and wavelet transformations. However, the combination of spectrogram and TSFEL features was proven to be most effective in highlighting the properties of the artifacts, leading to superior classification results. While usability detection can be accomplished using just one feature set (more details in Section 5.3), employing both sets enriches the model's understanding of the classes and enhances its robustness.

### 3.1.4. eegUsability: model training

We used the extracted features to train a LightGBM model [64,87]. LightGBM is a high-performance gradient-boosting framework designed for efficiency and scalability and is particularly well-suited for large datasets and complex ML tasks. While it shares similarities with traditional decision tree-based boosting methods, LightGBM distinguishes itself by adopting a leaf-wise growth strategy rather than the conventional level-wise approach. This method prioritizes the expansion of the leaf that offers the highest information gain, allowing for deeper tree development and improved performance with fewer iterations. Renowned for its rapid training speed, reduced memory consumption, and effective management of categorical features, LightGBM consistently outperforms other contemporary algorithms by delivering superior results [88].

LightGBM, like other gradient-boosting machines, attempts to minimize a loss function during model training. Let us represent the training dataset with $\boldsymbol{X_{tr} = F_{sp} + F_{ts}}$ (for all the samples of the training subset). LightGBM minimizes the categorical cross-entropy loss $L_o$, which is applicable to our multiclass classification problem and is expressed by:

$$L_o = -\sum_{i=1}^{N}\sum_{k=1}^{K} y_{ik} \log(p_{ik}) \qquad (4)$$

where $N$ is the number of data points, $K$ is the number of classes, $y_{ik}$ is a binary indicator (0 or 1) if class-label $k$ is the correct classification for observation $i$, and $p_{ik}$ is the predicted probability that observation $i$ belongs to class $k$. Probabilities are computed using the softmax function, which converts the logits $z_{ik}$ from the model output into normalized probabilities:

$$p_{ik} = \frac{\exp(z_{ik})}{\sum_{j=1}^{K} \exp(z_{ij})} \qquad (5)$$

As mentioned earlier, LightGBM builds trees leaf-wise rather than level-wise, optimizing the best leaf to split based on the highest gain from the gradient statistics and Hessians, which are defined as:

$$\text{Gradient}_{ik} = p_{ik} - y_{ik} \qquad (6)$$

$$\text{Hessian}_{ik} = p_{ik} \cdot (1 - p_{ik}) \qquad (7)$$

The model is then updated iteratively by adding a new tree that best fits the negative gradients:

$$m_{i+1}(x) = m_i(x) + \eta \cdot h_i(x) \tag{8}$$

where $m_i(x)$ represents the model prediction at iteration $i$, $\eta$ is the learning rate, and $h_i(x)$ is the output of the tree added at iteration $i$. Regularization is crucial to prevent overfitting and is implemented through feature and data subsampling. To build the eegUsability model, we used 80% of the features for individual tree building, 70% of $X_{tr}$ for building trees in every 10 iterations, and $\eta = 0.01$ to avoid overfitting.

### 3.1.5. eegUsability: performance evaluation

In this section, we present the performance of the eegUsability model on previously unseen samples from the test subset based on metrics widely used to assess SL models, namely, the confusion matrix, receiver operating characteristics (ROC) curves, precision, recall, F1-score, the area under curve (AUC), and κ [89]. A confusion matrix offers the exact counts (or percentages) of samples from each class that were correctly classified or misclassified and identifies the class to which misclassified samples are attributed. An ROC curve graphically represents the model's ability to correctly classify samples at various threshold levels by plotting the true positive rate against the false positive rate. Precision and recall, metrics derived from the confusion matrix, quantify the accuracy of the model in identifying true members of a class and the extent to which all relevant samples of a class are identified. The F1-score, a weighted mean of precision and recall, provides a singular measure to reflect the true performance of the classifier, which is particularly useful in contexts of an imbalanced sample ratio. The AUC measures the overall effectiveness of the classifier by calculating the area beneath the ROC curve, providing a comprehensive metric of performance across different class thresholds. Finally, κ quantifies the agreement level between two raters (or, in this case, a scorer and a classifier) in assigning categorical labels to a set of samples, adjusting for the probability of random agreement, which offers a more robust measure than simple accuracies or F1-scores. Apart from ROC curves, which visually represent AUC values, all these metrics range between 0 and 1 (or 0% and 100%, in percentage terms), except for κ, which ranges from −1 to 1. In all cases, a higher value indicates better performance of the classifier.

Figure 3 illustrates the performance of the eegUsability model using a confusion matrix, class-wise ROC curves, and AUC values. As depicted in Figure 3(a), the model achieves high recall scores across most classes, with notable success in recognizing Usable, No Data, and High Noise classes. Specifically, the classifier identifies usable data with an impressive 93.55% accuracy, a key takeaway for most studies. However, the M-shaped Noise poses a significant challenge, with 51.46% of its instances mistakenly classified as good data, highlighting difficulties in distinguishing between these categories. The classifier also effectively identifies Spiky Noise, though there is room for improvement. These observations are consistent with the class-wise ROC curves shown in Figure 3(b), where usable data nearly achieves a perfect AUC score, while M-shaped Noise records the lowest. Additional quantitative metrics such as precision, recall, and F1-score are detailed in Table 1 for each class and as a weighted mean, reflecting the proportion of each class in the test subset. As seen from the table, the eegUsability model demonstrates a weighted mean precision, recall, F1-score, and AUC of 84.98%, 86.01%, 84.87%, and 0.96, respectively. Furthermore, a κ value of 0.78 indicated substantial agreement between the two sets of scores. Several variants of the eegUsability model were developed as well, each potentially offering advantages depending on the application. These variants are discussed in Section 5.3, and their performance has been detailed in Appendix 2.

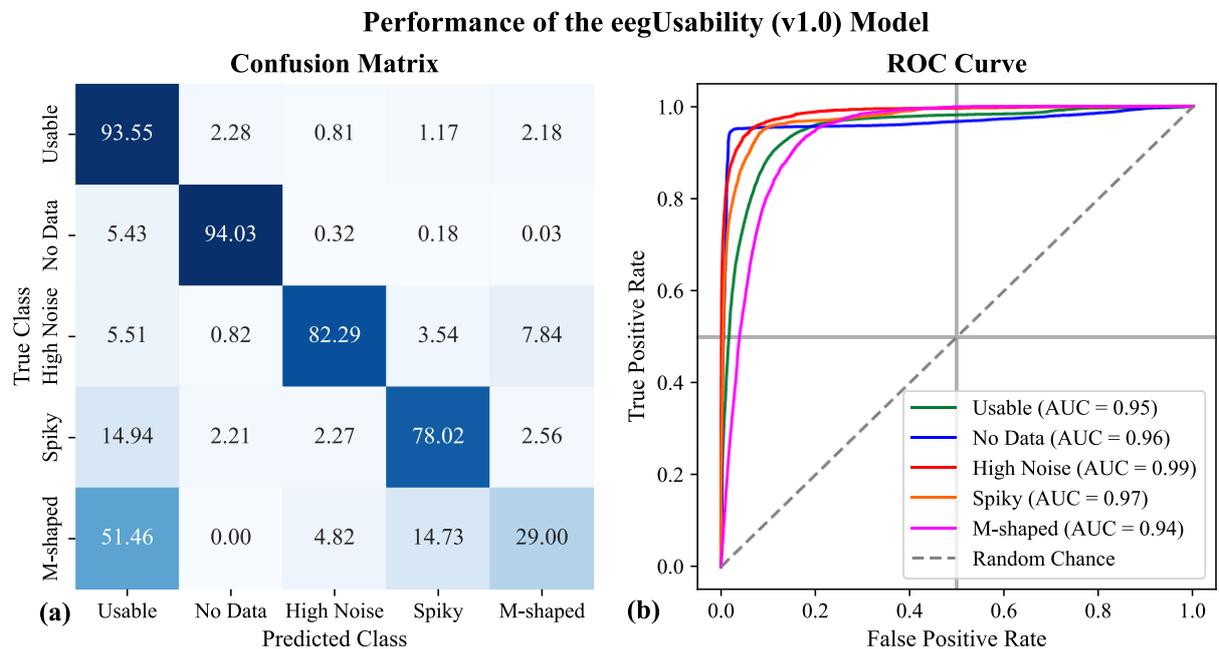

**Figure 3:** Class-wise performance of the eegUsability model, expressed in terms of **(a)** a confusion matrix (in percentages) and **(b)** ROC curves.

**Table 1:** Class-wise and aggregated performance of the eegUsability model.

| Class name | Precision (%) | Recall (%) | F1-score (%) | AUC | Test samples (class ratio, %) |
|:---:|:---:|:---:|:---:|:---:|:---:|
| Usable | 86.06 | 93.55 | 89.65 | 0.95 | 26894 (50) |
| No Data | 95.25 | 94.03 | 94.64 | 0.96 | 15186 (28.23) |
| High Noise | 83.46 | 82.29 | 82.87 | 0.99 | 3557 (6.61) |
| Spiky | 66.86 | 78.02 | 72.01 | 0.97 | 3126 (5.81) |
| M-Shaped | 60.53 | 29 | 39.21 | 0.94 | 5025 (9.34) |
| **Weighted Mean:** | **84.98** | **86.01** | **84.87** | **0.96** | **Total: 53788** |

## 3.2. Model eegMobility

eegMobility, the second core model of the eegFloss package, is another SL model designed to analyze the degree of mobility from Zmax ACC data associated with its EEG channels. The output of this model is used to detect time-in-bed (TIB) automatically, a key measure in sleep research. It can also be used to truncate overnight EEG recordings to remove the parts not related to sleep (especially from the beginning and the end). TIB refers to the total duration an individual spends in bed with the intention of sleeping, which spans from the moment they lie down (at night) to the moment they get up (in the morning), irrespective of the actual sleep period [90]. Traditional PSG does not typically include ACC. Hence, most sleep studies assume that the entire duration of the recording equates to TIB during their analysis, which can be misleading, as individuals may not go to bed or fall asleep immediately after starting the recording and may not turn off the device immediately upon waking up. Fortunately, ACC sensors are a part of most modern wearables and PSG systems, which allows detecting when someone lies down and gets up more precisely by utilizing human activity recognition (HAR) models [91]. In eegFloss, we denote them as *Lights Out* and *Lights On* moments, respectively. We address the automatic TIB detection in two stages: initially, we employ the eegMobility model to assess the degree of movement from the tri-axial ACC signals. Then, we use its classification outcomes (*mobility scores*) combined with a predefined threshold to identify the Lights Out and Lights On moments, which are the boundaries of TIB. Below, we outline the formulation of the eegMobility model, which employs a methodology similar to that of the eegUsability model discussed in Section 3.1.

### 3.2.1. eegMobility: data collection

The eegMobility model is trained on a relatively small (≈23 hours in total) seeded dataset of ACC data collected using the Zmax headband while performing eight daily activities, namely lying (including two full nights of sleep), sitting, standing, walking, ascending and descending a set of stairs, jumping, and biking. However, since we were only interested in identifying the periods of (uninterrupted) lying, precise differentiation among the other activities was not essential. Hence, we categorized the other seven activities into two groups based on the degree of movement involved: *Stationary* (sitting and standing) and *Mobile* (the remaining five activities). Additionally, we observed that participants sometimes forget to turn off the recording after waking up (and detaching the device from their forehead). This scenario is accounted for and labeled as *Idle*—the device is not being used but continues to record data. Therefore, we could define this as a four-class classification problem with the class labels *Lying*, *Stationary*, *Mobile*, and *Idle*.

It is important to note that *Lying* in the training dataset encompasses various movements performed while the participant is in bed, including actions such as changing sides, reading, and using a smartphone, to better reflect real-world conditions. As per the definition, TIB begins when the participant gets into bed and ends when they get out of it, regardless of the activities performed during that period. Accordingly, we aimed to home in on these moments by taking into account not only the magnitude of movement detected by the tri-axial ACC, but the directional changes across its axes as well.

### 3.2.2. eegMobility: feature extraction

Samples were derived from the raw tri-axial Zmax ACC data. To ensure proper alignment with usability scores (later on), the window size to prepare samples to train this model was set to 10 seconds as well. Consequently, each sample of the eegMobility model comprised three ($A_x$, $A_y$, and $A_z$) 10-second signal segments. In the resultant set of samples, the class proportions of the four mentioned classes were approximately 57.27%, 16.45%, 21.68%, and 4.6%, making it an imbalanced dataset. However, due to the distinct characteristics of each class, we hypothesized that the imbalance would not pose a significant challenge. After preparing the samples, we extracted TSFEL features from them. As outlined in Section 3.1.3.2, the library produced 390 features from each signal segment, resulting in 1170 features per sample.

### 3.2.3. eegMobility: training and performance

eegMobility is also a LightGBM-based model, and its training process was similar to that of eegUsability, except for a few key changes. eegMobility was trained only on ACC data (with fewer features per sample). Here, we implemented a random 80:20 split to allocate the dataset into training and test sets, carefully maintaining the original class ratios to mitigate the effects of class imbalance. Additionally, specific LightGBM parameters were fine-tuned to optimize the model's performance.

Figure 4 displays the eegMobility model's performance on the test subset through a confusion matrix, class-wise ROC curves, and AUC values. The results indicate near-perfect scores across all classes, with a perfect recall for Lying—our primary focus. These results are corroborated by the performance metrics detailed in Table 2. Furthermore, the model achieved a κ score of 99.38, indicating an almost perfect agreement with the true labels, suggesting class imbalance did not significantly impact the classifier's effectiveness. A "lite" version of the model, which operates solely on power features (instead of TSFEL features) extracted using Welch's method [92], is also available that offers competitive performance while significantly reducing processing time (detailed in Appendix 2).

HAR is a well-studied area, and it is common for HAR models to exhibit excellent classification performance, especially when the activities are distinctly different, even with limited data and smaller window sizes [93,94]. HAR models are typically trained on samples ranging from 2.5 to 3 seconds each [95]. In our study, employing a considerably larger window size may have contributed to the model's near-perfect classification scores. Although testing with shorter window sizes is possible, the standard epoch length in sleep research is 30 seconds, making even our 10-second epochs relatively brief. Since the primary objective of eegMobility is to facilitate automatic TIB detection to refine autoscoring, a shorter window size would likely offer little additional benefit for further analysis. Identification of the Lights Out and Lights On moments and TIB from the obtained mobility scores have been elaborated in Section 4.3.

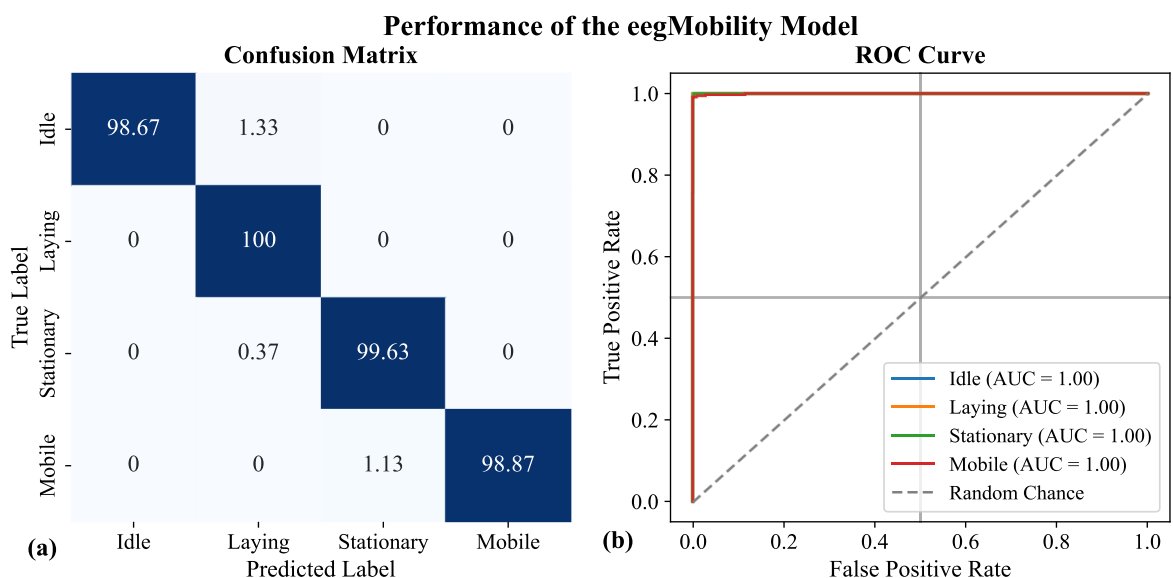

**Figure 4:** Class-wise performance of the eegMobility model, expressed in terms of **(a)** a confusion matrix (in percentages) and **(b)** ROC curves.

Table 2: Class-wise and aggregated performance of the eegMobility model.

| Class name | Precision (%) | Recall (%) | F1-score (%) | AUC | Test samples (class ratio, %) |
|---|---|---|---|---|---|
| Idle | 100 | 98.67 | 99.33 | 1 | 75 (4.60) |
| Lying | 99.79 | 100 | 99.89 | 1 | 934 (57.27) |
| Stationary | 98.52 | 99.63 | 99.07 | 1 | 268 (16.43) |
| Mobile | 100 | 98.87 | 99.43 | 1 | 354 (21.70) |
| **Weighted Mean:** | **99.64** | **99.63** | **99.63** | **1** | **Total: 1631** |

## 4. eegFloss: package functionalities

eegFloss is a Python package containing a set of scripts to apply the eegUsability model to assess the epoch-wise usability of sleep EEG data from various devices and the eegMobility model to evaluate the degree of mobility from Zmax ACC data. Additionally, it can aggregate existing sleep scores with usability scores to produce artifact-rejected scores, generate hypnograms, filter out Spiky Noise from EEG data, and compute detailed sleep statistics. To ensure optimal functionality, the package should be executed within a specifically configured Python environment, which includes all necessary packages. Detailed instructions for setting up the environment and utilizing the package are provided in a Markdown (MD) file named *README.MD*, located in the associated code repository (refer to the Code availability section). In this section, we describe these functionalities and their inner workings in detail, which also serves as a guide for using the package. A flowchart of the workflow of the package's primary script is presented in Figure 5.

### 4.1. Data usability detection

For a given EEG recording, first, eegFloss segments each EEG channel's data into 10s-epochs. Next, if parallel tri-axial ACC data is available, it pairs two signal segments to create a sample—one containing raw EEG and the other containing $A_{norm}$. Then, it extracts two sets of features from each paired sample ($F_{sp}$ and $F_{ts}$) and concatenates them horizontally to form a set of 3618 features per sample, as detailed in Section 3.1.3. After that, it employs the eegUsability model to determine the presence of artifacts in each sample based on the extracted features. This classification produces a set of usability labels ($U$) for each EEG channel, indicating whether the data is suitable for autoscoring and, if not, identifying which artifacts (described in Section 3.1.1) are present. Finally, it exports a graphical representation of the channel-wise usability scores (called a *usability graph*), providing a means to (visually) verify the eegUsability outcomes. Figure 6 presents an example usability graph of a sample Zmax recording. Refer to the Supplementary materials section to see usability graphs of over 900 nights from five Zmax datasets generated by eegFloss.

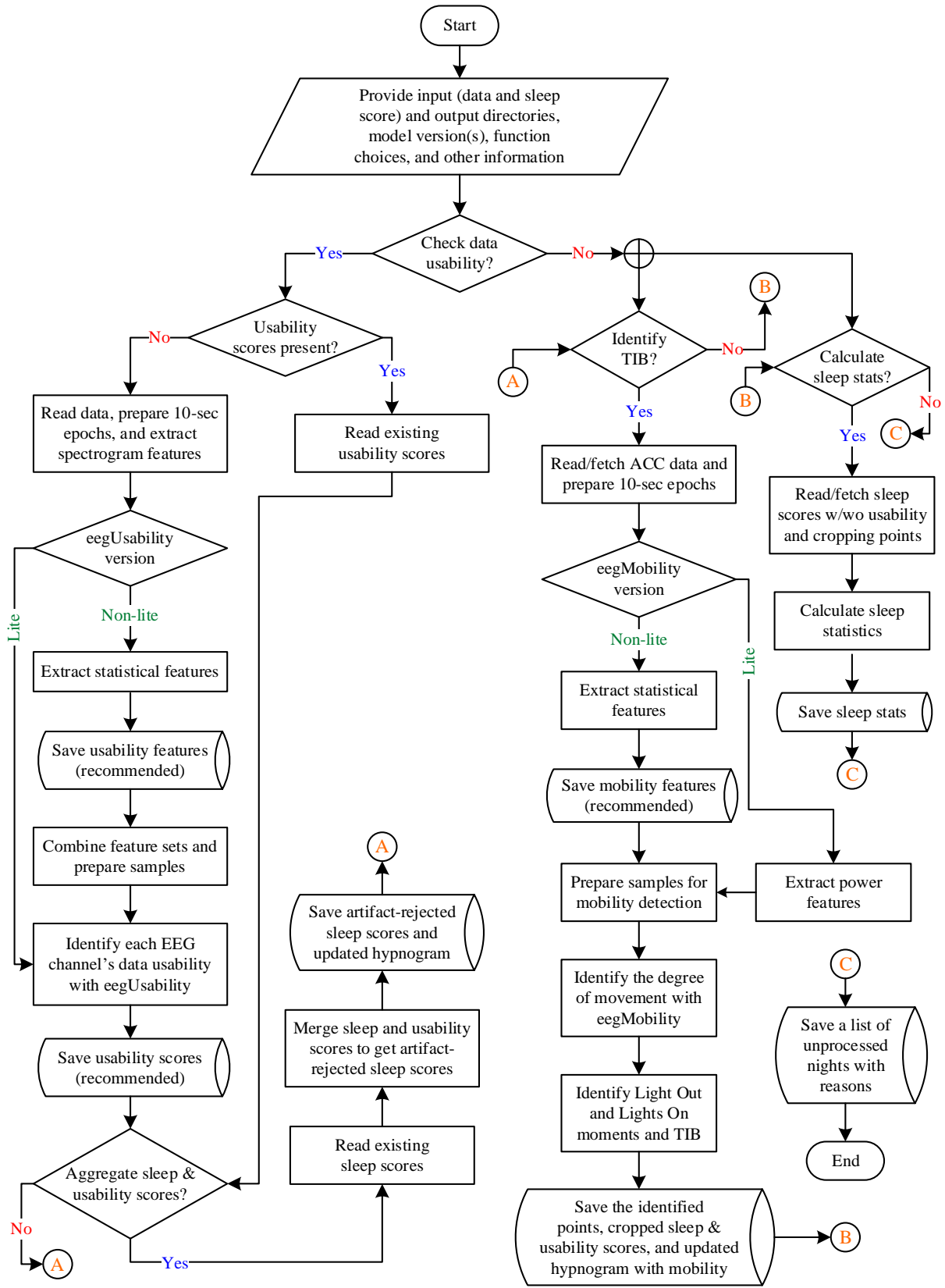

**Figure 5:** A simplified workflow of the eegFloss package's primary script.

## 4.2. Obtaining artifact-rejected scores

eegFloss currently does not feature a native autoscorer. Therefore, sleep scores must be provided alongside the EEG data for the aggregation process to commence. Typically, sleep scoring is performed in 30-second epochs, and the resultant autoscores ($S_s$) can be represented as $S_{s,i} \in \{W, N1, N2, N3, REM\}$. However, eegFloss accommodates any epoch length of sleep scoring that is a direct multiple of the window size used for usability detection. This flexibility is essential for generating artifact-rejected scores. Additionally, the default window size for usability detection, which is set at 10 seconds, can be adjusted if necessary (more details in Section 5.2). However, to ensure flawless execution of the package under these variations, specific parameters within eegFloss may require adjustments.

Once the autoscores and usability scores (of a given night) are obtained, eegFloss implements a three-step aggregation process to obtain the artifact-rejected scores. First, channel-wise usability scores are binarized such that "0" indicates usable (artifact-free) data, and a non-zero label denotes an artifact (regardless of the type). For a device with $n$ EEG channels, let the channel-wise usability scores be denoted as $U_c$, where $c = 1, 2, ..., n$ and $U_{c,i} \in \{0, 1, 2, 3, 4\}$. The binarized scores ($B_c$) are defined as:

$$B_{c,i} = \begin{cases} 0 & \text{if } U_{c,i} = 0 \\ 1 & \text{if } U_{c,i} \neq 0 \end{cases} \quad (9)$$

where 0 and 1 represent usable and unusable epochs, respectively. These binarized scores are aggregated across channels using a *majority rule*—if the majority of channels in an epoch are marked as unusable, the epoch is considered unusable. The aggregated usability scores ($U_{agg}$) can be expressed as:

$$U_{agg,i} = \begin{cases} 1 & \text{if } \frac{\sum_{c=1}^{n} B_{c,i}}{n} > 0.5 \\ 0 & \text{otherwise} \end{cases} \quad (10)$$

Then, since sleep and usability scores (usually) differ in epoch length, usability scores are further aggregated to match the sleep epoch resolution. This is done by grouping consecutive epochs based on a *scaling factor* ($s_f$), which defines how many shorter usability epochs fit into a single sleep epoch. The second aggregation also follows a majority rule—if the majority of the epochs within $s_f$ are unusable, the aggregated epoch is considered unusable:

$$U_{s_f,i} = \begin{cases} 1 & \text{if } \frac{\sum_{j=i \cdot s_f+1}^{(i+1) \cdot s_f} U_{agg,j}}{s_f} > 0.5 \\ 0 & \text{otherwise} \end{cases} \quad (11)$$

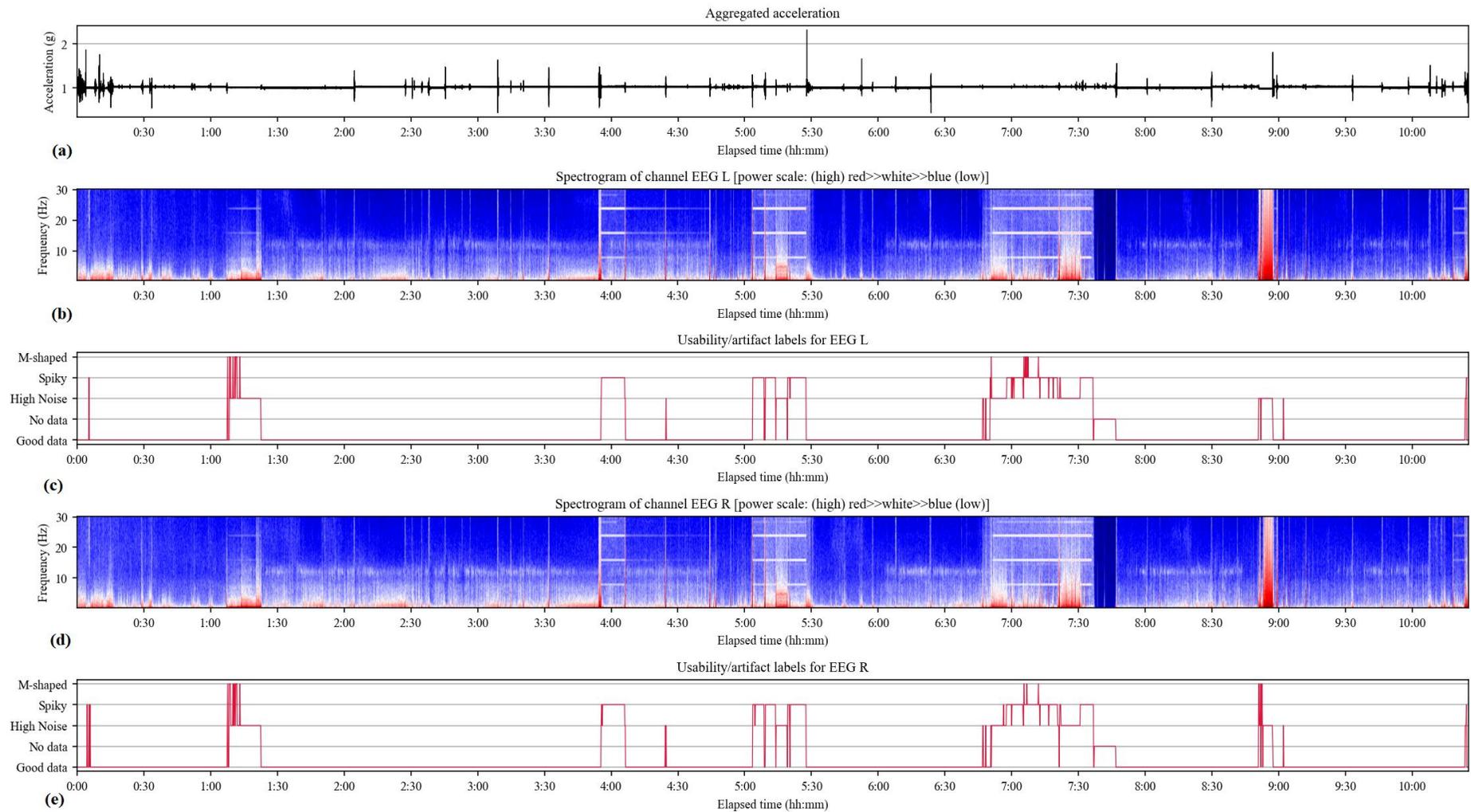

**Figure 6:** The usability graph of a sample Zmax recording showing **(a)** the normalized acceleration calculated from tri-axial ACC data, **(b)** a windowed spectrogram of the EEG Left channel, **(c)** its usability scores, **(d)** a windowed spectrogram of the EEG Right channel, and **(e)** its usability scores.

Finally, the artifact-rejected sleep scores ($S_{ar}$) are obtained by combining the aggregated usability scores ($U_{s_f}$) with the original sleep scores ($S_s$). If an epoch is deemed unusable, it is marked as such; otherwise, the original sleep stage is retained:

$$S_{ar,i} = \begin{cases} un & \text{if } U_{sf,i} = 1 \\ S_{s,i} & \text{otherwise} \end{cases} \quad (12)$$

where $un$ represents an unscorable epoch. While equations (10–12) describe the default behavior of the eegFloss package, these can be modified according to the needs of a specific project. By default, eegFloss expects $S_{s,i} \in \{W, N1, N2, N3, REM\} \rightarrow \{0, 1, 2, 3, 4/5\}$ and uses "−1" to indicate unscorable epochs, meaning:

$$S_{ar,i} \in \{un, W, N1, N2, N3, REM\} \rightarrow \{-1, 0, 1, 2, 3, 4\} \quad (13)$$

Figure 7 visually represents the three-step aggregation process implemented by eegFloss for Zmax data, where $U_L$ and $U_R$ represent EEG Left and Right channels, respectively.

| $U_L$ : | 0 | ≠0 | 0 | ≠0 | ≠0 | 0 | ≠0 | 0 | ≠0 | ≠0 | ≠0 | ≠0 | ≠0 | ≠0 | ≠0 | ≠0 | 0 | 0 |
|---|---|---|---|---|---|---|---|---|---|---|---|---|---|---|---|---|---|---|
| $U_R$ : | 0 | 0 | 0 | 0 | ≠0 | 0 | ≠0 | 0 | ≠0 | ≠0 | ≠0 | ≠0 | ≠0 | ≠0 | 0 | ≠0 | ≠0 | ≠0 |
| $U_{agg}$ : | 0 | 0 | 0 | 0 | 1 | 0 | 1 | 0 | 1 | 1 | 1 | 1 | 1 | 1 | 0 | 1 | 0 | 0 |
| $U_{s_f}$ : | 0 | | | 0 | | | 1 | | | 1 | | | 1 | | | 0 | | |
| $S_s$ : | W | | | N1 | | | N2 | | | N3 | | | REM | | | REM | | |
| $S_{ar}$ : | W | | | N1 | | | $un$ | | | $un$ | | | $un$ | | | REM | | |

**Figure 7:** A visual representation of the three-step aggregation process employed by eegFloss to obtain artifact-rejected scores from sleep and usability scores.

## 4.3. Determining TIB

eegFloss can automatically detect TIB from mobility scores provided by the eegMobility model. This model is trained on Zmax ACC data and is not recommended for use on ACC data from other devices, as misclassification may occur due to variations in device positioning and the axis directions of the tri-axial ACC sensors. For a given Zmax recording, if we denote a set of mobility scores as $M_b$, where $M_{b,i} \in \{\text{Idle, Lying, Stationary, Mobile}\}$, eegFloss defines the Lights Out and Lights On moments (in seconds) as:

$$\text{Lights Out} = 10 \times \min\{i \mid M_{b,i}, M_{b,i+1}, \ldots, M_{b,i+11} = \text{Lying}\} \quad (14)$$

$$\text{Lights On} = 10 \times \max\{i + 11 \mid M_{b,i}, M_{b,i+1}, \ldots, M_{b,i+11} = \text{Lying}\} \quad (15)$$

where $1 \leq i \leq |M_b| - 11$. In other words, the Lights Out moment is defined as the first second of the initial occurrence of two minutes of consecutive Lying, while the Lights On moment marks the last second of the last occurrence of two minutes of consecutive Lying. However, the two-minute threshold is not fixed; alternative thresholds can be easily implemented within the eegFloss package. Thus, TIB (in minutes) can be determined using:

$$\text{TIB} = \frac{\text{Lights On} - \text{Lights Out} + 1}{60} \qquad (16)$$

## 4.4. Creating hypnograms

eegFloss can generate hypnograms upon data processing, depicting different sets of information. Figure 8 demonstrates a sample hypnogram drawn by eegFloss, illustrating the results of the processing steps described in previous subsections. Specifically, Figure 8(d) displays $S_{ar}$ obtained from Section 4.2, while Figure 8(e) depicts $M_b$, the Lights Out and Lights On moments, and TIB as described in Section 4.3. These visual representations may help in comprehending the outcomes from different processing stages and identifying any significant misclassifications or irregularities. For additional examples, refer to the Supplementary materials section, which includes hypnograms of over 900+ nights from five Zmax datasets generated by eegFloss.

## 4.5. Filtering out the spiky noise

The Spiky Noise (described in Section 3.1.1.1) differs from the other artifacts found in the Zmax data. It appears to be often superimposed on raw EEG data, which may otherwise be good data or contain other artifacts. Therefore, filtering out the spiky noise epoch may yield usable data. As demonstrated in the windowed spectrogram shown in Figure 2(a), Spiky Noise is identified by its high power components at ≈8 Hz and its harmonic frequencies. To effectively eliminate this noise, eegFloss employs a Butterworth lowpass filter combined with a cascaded Notch filter [96]. The sequential operations of these filters are described below.

For a segment of EEG data ($f_s = 256$ Hz) containing Spiky Noise, eegFloss first filters the signal segment $x(t)$ using a fourth-order Butterworth low-pass with a cutoff frequency $f_c$ of 30 Hz to remove all the high-frequency components from the signal. This is a widely adopted practice in sleep research, as sleep-related information primarily resides below this threshold [97]. The normalized cutoff frequency ($f'_c$) of the Butterworth filter can be calculated using:

$$f'_c = \frac{f_c}{f_s/2} \qquad (17)$$

where the frequency is normalized by the Nyquist frequency ($f_s/2$). The transfer function in the Laplace domain $H_B(s)$ can be expressed as:

$$H_B(s) = \frac{1}{\prod_{k=1}^{n}(s - s_k)} \qquad (18)$$

where $s_k$ represents the poles of the Butterworth polynomial and is defined as:

$$s_k = \omega_c e^{j\frac{(2k+n-1)\pi}{2n}} \qquad (19)$$

where $\omega_c$ is the cutoff angular frequency of the Butterworth filter and $k = 1, 2, ..., n$.

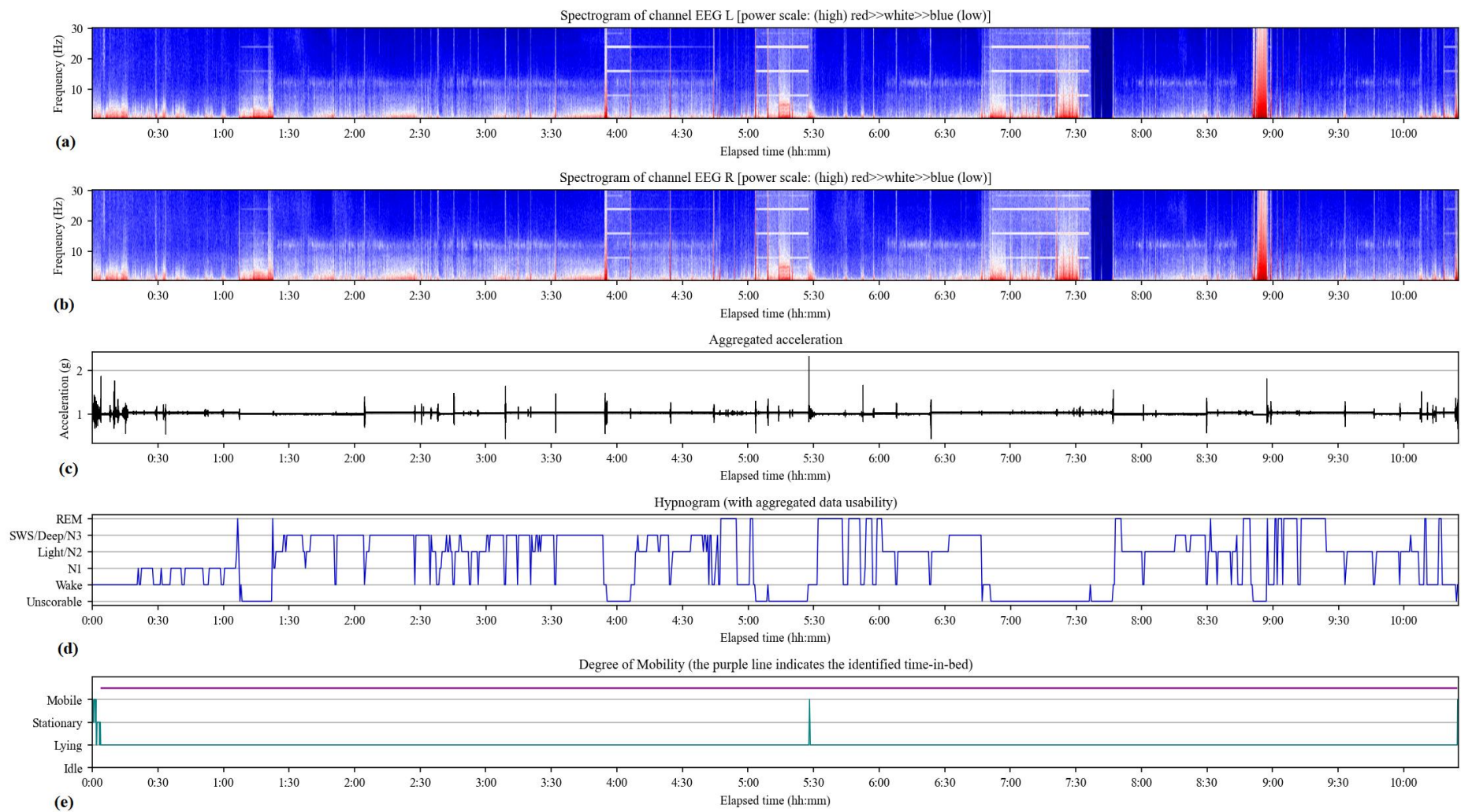

**Figure 8:** eegFloss outputs of a sample Zmax recording showing spectrograms of **(a)** EEG Left and **(b)** EEG Right channels, **(c)** the normalized acceleration, **(d)** hypnogram based on the artifact-rejected Dreamento autoscores, and **(e)** the mobility labels with TIB bounded by Lights Out and Lights On moments.

The bilinear transformation is used to map from the Laplace domain to the Z-domain:

$$s = \frac{2}{T_s} \frac{1 - z^{-1}}{1 + z^{-1}} \tag{20}$$

The Butterworth filter's transfer function in the Z-domain $H_B(z)$ can be expressed as:

$$H_B(z) = \frac{B(z)}{A(z)} = \frac{b_0 + b_1 z^{-1} + ... + b_n z^{-n}}{a_0 + a_1 z^{-1} + ... + a_n z^{-n}} \tag{21}$$

where $A(z)$ and $B(z)$ represent the coefficients of the filter's numerator and denominator, respectively, and $a_k$ and $b_k$ are the filter coefficients. The filtered signal $y_{bt}(t)$ is obtained using zero-phase filtering:

$$y_{bt}(t) = \sum_{k=0}^{n} b_k\, x(t - kT_s) - \sum_{k=1}^{n} a_k\, y_{bt}(t - kT_s) \tag{22}$$

where $T_s = 1/f_s$ is the sampling period. After low-pass filtering, eegFloss applies a cascaded Notch filter (at 8 Hz, 16 Hz, and 24 Hz) with a bandwidth of 2 to remove periodic noise components from $y_{bt}(t)$. The normalized frequency $\omega_o$ and bandwidth $bw$ of a Notch filter can be calculated using:

$$\omega_o = \frac{f_n}{f_s/2} \tag{23}$$

$$bw = \frac{2}{f_s/2} \tag{24}$$

where $bw$ is related to the quality factor ($Q = \omega_o/bw$). The transfer function of a second-order notch filter is given by:

$$H_N(s) = \frac{s^2 + \omega_o^2}{s^2 + 2r\omega_o s + \omega_o^2} \tag{25}$$

where $r = 1 - (bw/2)$ controls the notch width. In the Z-domain, the transfer function of the notch filter can be expressed as:

$$H_N(z) = \frac{1 - 2\cos(2\pi\omega_o)z^{-1} + z^{-2}}{1 - 2r\cos(2\pi\omega_o)z^{-1} + r^2 z^{-2}} \tag{26}$$

For each noise frequency $f_n$, a notch filter $H_{N_f}(z)$ is designed, and the overall filter response is obtained by cascading all notch filters with the Butterworth filter:

$$H_{cs}(z) = H_B(z) \prod_{f_n} H_{N_f}(z) \tag{27}$$

This is implemented by convolving the filter coefficients:

$$B_{cs}(z) = B_B(z) * B_N(z) \tag{28}$$
$$A_{cs}(z) = A_B(z) * A_N(z) \tag{29}$$

where $*$ denotes convolution. Finally, the filtered signal $y_{cs}(t)$ using the cascaded filter coefficients $a_{cs}$ and $b_{cs}$ can be expressed as:

$$y_{cs}(t) = \sum_{k=0}^{n} b_{cs,k}\, x(t - kT_s) - \sum_{k=1}^{n} a_{cs,k}\, y_{cs}(t - kT_s) \qquad (30)$$

Figure 9 presents a sample Zmax EEG signal $x(t)$, along with the filtered signals $y_{bt}(t)$ and $y_{cs}(t)$ obtained after applying the described Butterworth and cascaded filters, respectively. As observed in the figure, the filtering process effectively removes most of the noise components, resulting in a signal that closely resembles a clean sleep EEG signal. Figure 10 displays the frequency response of the Spiky Noise filter described above.

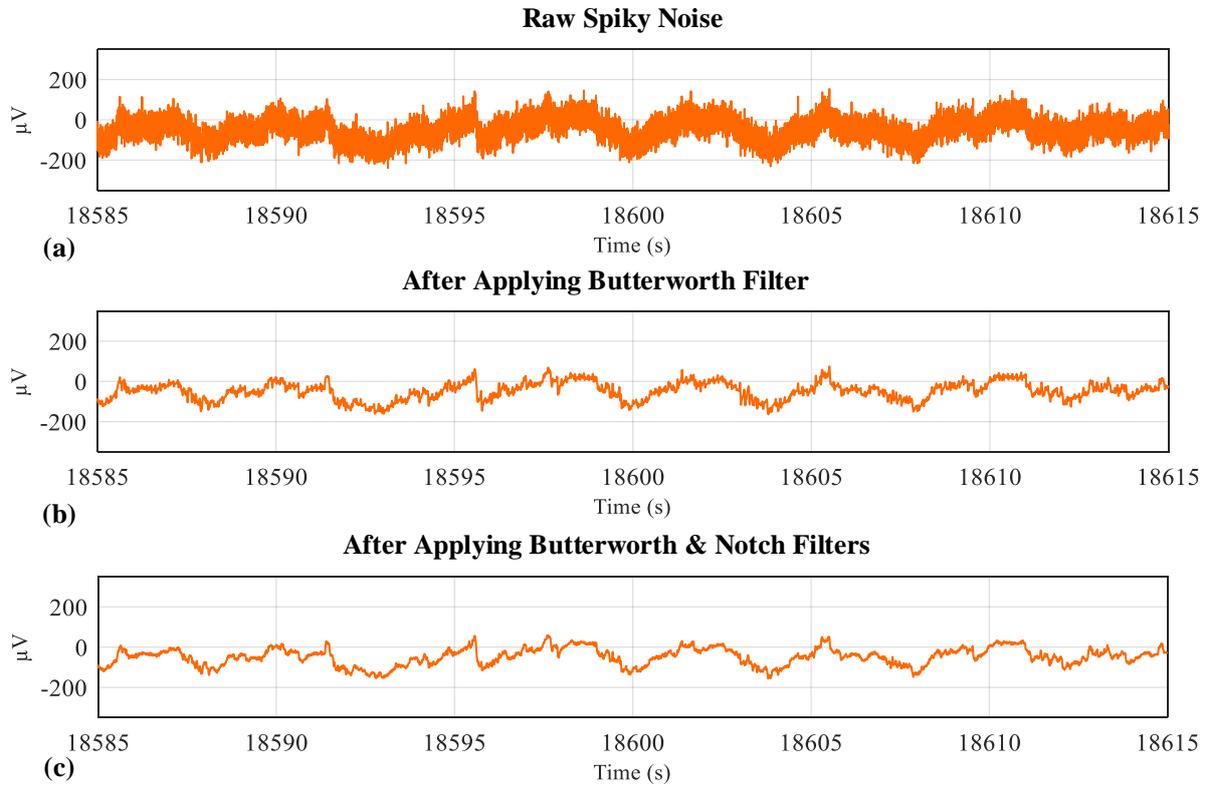

**Figure 9:** **(a)** A sample Zmax EEG segment contaminated with Spiky Noise, **(b)** the signal after applying a Butterworth low-pass filter, and **(c)** the signal after the removal of Spiky Noise using the (Butterworth-Notch) cascaded filter.

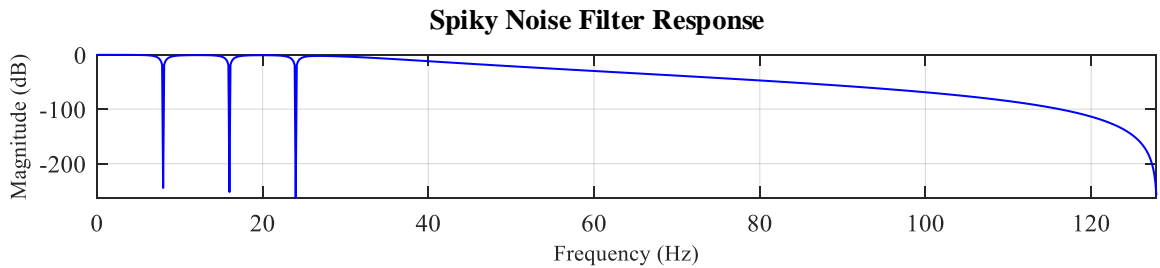

**Figure 10:** The frequency response of the Butterworth-Notch cascaded filter used for filtering out the Spiky Noise.

## 4.6. Calculating sleep statistics

eegFloss can calculate standard sleep metrics from $S_{ar}$ or $S_s$, as outlined and detailed in Table 3. These metrics are well-defined in sleep literature [25,98]. Although their formulation was inspired by the *YASA sleepstats* module [99], the calculations were modified in eegFloss to account for unscorable epochs and maintain compatibility with the other processing steps. In addition to these metrics, the Lights Out and Lights On moments (in seconds) are also incorporated into the eegFloss's sleep statistics module outputs.

## 4.7. Checking the usability of non-Zmax EEG data

Although eegUsability is trained on Zmax data, it can be applied to sleep EEG data from other devices using the eegFloss package. The procedure for generating usability scores is consistent with that described for Zmax data in Section 4.1. Two primary factors that can influence the performance of eegUsability on non-Zmax data are the availability of raw tri-axial ACC outputs and the amplitude range of EEG signals during normal sleep. While ACC is common in most wearables, some sleep EEG systems (such as SleepLoop and Brain Quick Plus Evolution PSG system [100]) still lack it, and some systems do not store the raw tri-axial outputs (such as *SOMNOscreen plus* [101]). Although eegUsability can work without ACC input, its absence may lead to some epochs with arousals being classified as High Noise. Variations in referencing and the properties of the internal amplifiers and filters across devices can also affect the amplitude range of EEG signals. For instance, the amplitude range for usable data was observed to be within approximately ±100 µV in Zmax. The model may internalize this range when assessing data usability and provide erroneous outputs if it is not maintained. Therefore, specific normalization techniques might be necessary to mitigate these issues by examining the properties of (usable) data from new devices. For reference, the Zmax EEG data used to train (and evaluate) the eegUsability model has a minimum and maximum amplitude of −1976 µV and 1975.93 µV, respectively (including artifacts), and the normalized ACC data ranges from 0.018 g to 3.464 g and is centered around 1 g as shown in Figure 8(c). That said, in accordance with the AASM guidelines and standard PSG practices, artifact-free sleep EEG signals from any device are generally expected to fall within approximately ±100 µV. Segments exceeding these bounds are indicative of artifacts or non-physiological activity.

To demonstrate eegFloss's applicability on non-Zmax data, we utilized the Bitbrain Open Access Sleep (BOAS) dataset [102], which comprises 128 nights of simultaneous recordings from the Brain Quick Plus Evolution PSG system and the Bitbrain Ikon EEG headband [103,104]. Additionally, we applied the model to EEG data collected with the SOMNOscreen plus PSG system from the Wearanize+ dataset [30]. Figures 11, 12, and 13 present the usability graphs for three example nights recorded using the Brain Quick Plus Evolution PSG system, the SOMNOscreen Plus PSG system, and the Bitbrain Ikon EEG headband, respectively. As evident from these figures, eegUsability was largely effective in identifying the usable segments, which will be the primary focus of most studies. The detected No Data, High Noise, and Spiky Noise segments also align well with their definitions. The *M-shaped noise* is rarely observed in non-Zmax recordings, which suggests that it may be specific to Zmax.

**Table 3:** Sleep metrics and their definitions provided by eegFloss.

| Sleep metric | Metric name | Definition and calculation |
|---|---|---|
| Lights Out | Lights_out_sec | See Section 4.3, equation (14). |
| Lights On | Lights_on_sec | See Section 4.3, equation (15). |
| Scorable data (%) | Scorable_% | $\%\text{Scorable} = \left( \dfrac{\sum_{i=1}^{|S_{ar}|} \mathbf{1}[S_{ar,i} \neq -1]}{|S_{ar}|} \right) \times 100$ |
| Time-in-bed[*] | TIB_min | Equation (16), if Lights Out and Lights On moments are available; $\text{TIB} = |S_{ar}|/60$, if not. |
| Sleep Period Time[*] | SPT_min | The time span (in minutes) between the first and last non-wake epochs. |
| Total Sleep Time | TST_min | The total duration (in minutes) spent asleep during SPT. |
| N1 Sleep Time | N1_min | Total time (in minutes) spent in N1 sleep. |
| N1 Sleep (%) | N1_% | The percentage of N1 sleep during TST. |
| N2 Sleep Time | N2_min | Total time (in minutes) spent in N2 sleep. |
| N2 Sleep (%) | N2_% | The percentage of N2 sleep during TST. |
| N3 Sleep Time | N3_min | Total time (in minutes) spent in N3 sleep. |
| N3 Sleep (%) | N3_% | The percentage of N3 sleep during TST. |
| REM Sleep Time | REM_min | Total time (in minutes) spent in REM sleep. |
| REM Sleep (%) | REM_% | The percentage of REM sleep during TST. |
| NREM Sleep Time | NREM_min | Total time (in minutes) spent in non-REM sleep. |
| NREM Sleep (%) | NREM_% | The percentage of NREM sleep during TST. |
| Wake After Sleep Onset | WASO_min | Total duration (in minutes) spent awake between the first and last non-W epochs. |
| Sleep Onset Latency[*] | SOL_min | The time span (in minutes) between the Lights Out moment and the first detected non-W epoch. |
| N1 Sleep Latency[*] | N1_latency_min | The period between the Lights Out moment and the first N1 epoch (in minutes). |
| N2 Sleep Latency[*] | N2_latency_min | The period between the Lights Out moment and the first N2 epoch (in minutes). |
| N3 Sleep Latency[*] | N3_latency_min | The period between the Lights Out moment and the first N3 epoch (in minutes). |
| REM Sleep Latency[*] | REM_latency_min | The period between the Lights Out moment and the first REM epoch (in minutes). |
| Post Sleep Wakefulness | PSW_min | The time (in minutes) spent in bed after waking up before leaving the bed. |
| Sleep Efficiency | SE_% | The percentage of TST relative to TIB. |
| Sleep Maintenance Efficiency | SME_% | The percentage of TST relative to SPT. |

[*]Unscorable epochs are included in the calculation.

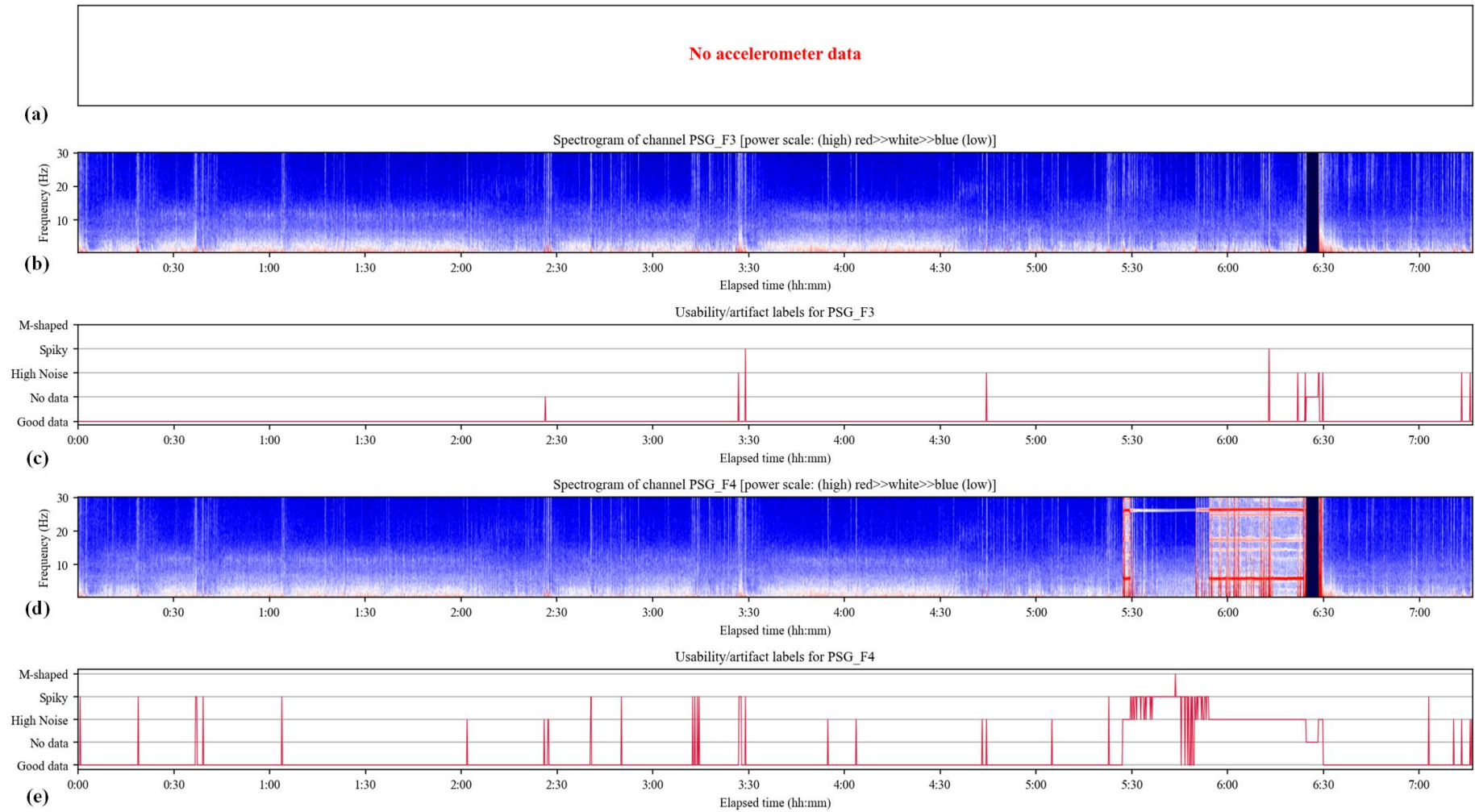

**Figure 11:** Usability graph of a Brain Quick Plus Evolution (PSG system) recording showing **(a)** normalized ACC outputs (unavailable for this device), **(b)** windowed spectrograms of the PSG_F3 channel, **(c)** usability scores for PSG_F3**, (d)** windowed spectrograms of the PSG_F4 channel, and **(e)** usability scores for PSG_F4.

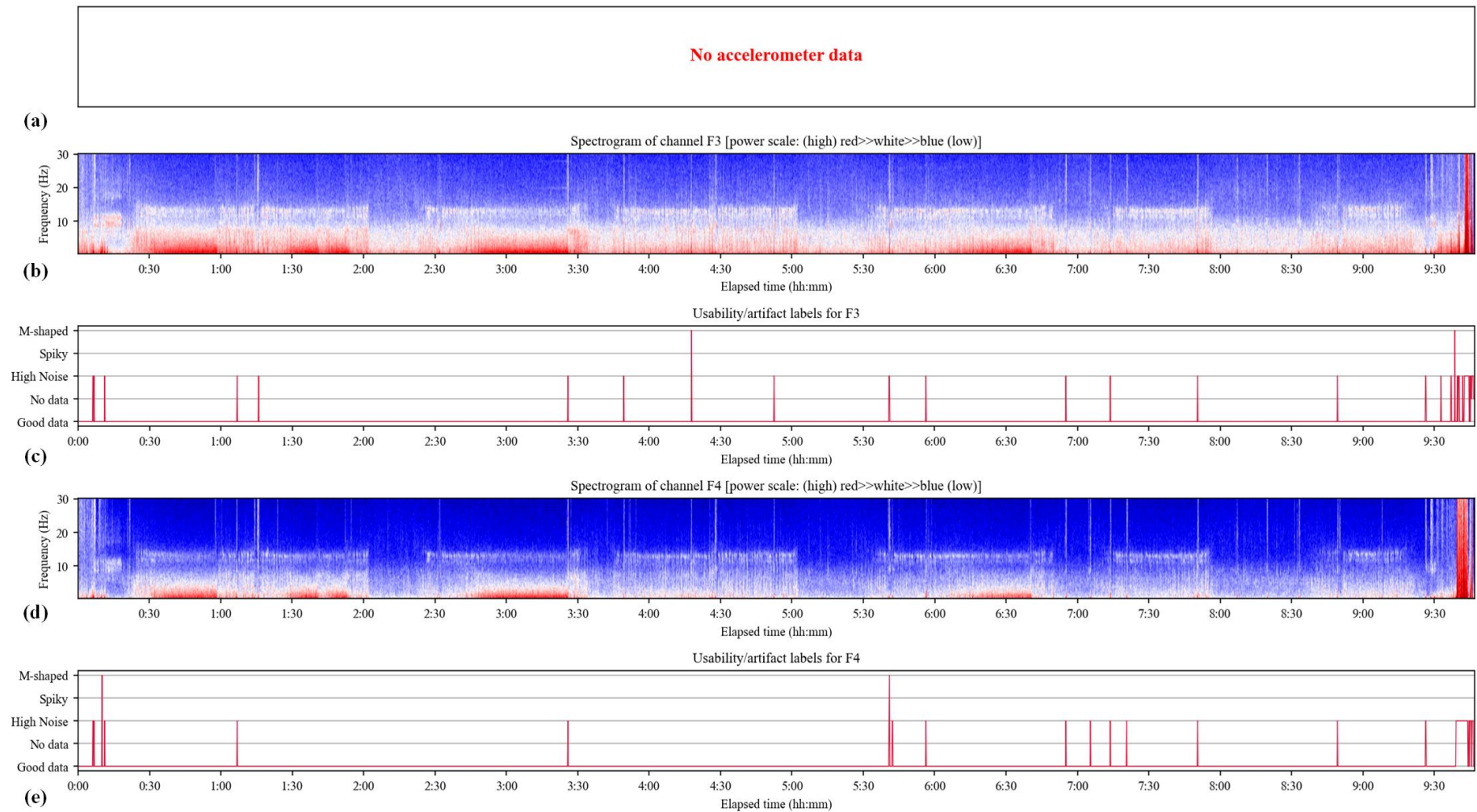

**Figure 12:** Usability graph of a SOMNOscreen plus (PSG system) recording showing **(a)** normalized ACC outputs (unavailable for this device), **(b)** windowed spectrograms of the F3 channel, **(c)** usability scores for F3, **(d)** windowed spectrograms of the F4 channel, and **(e)** usability scores for F4.

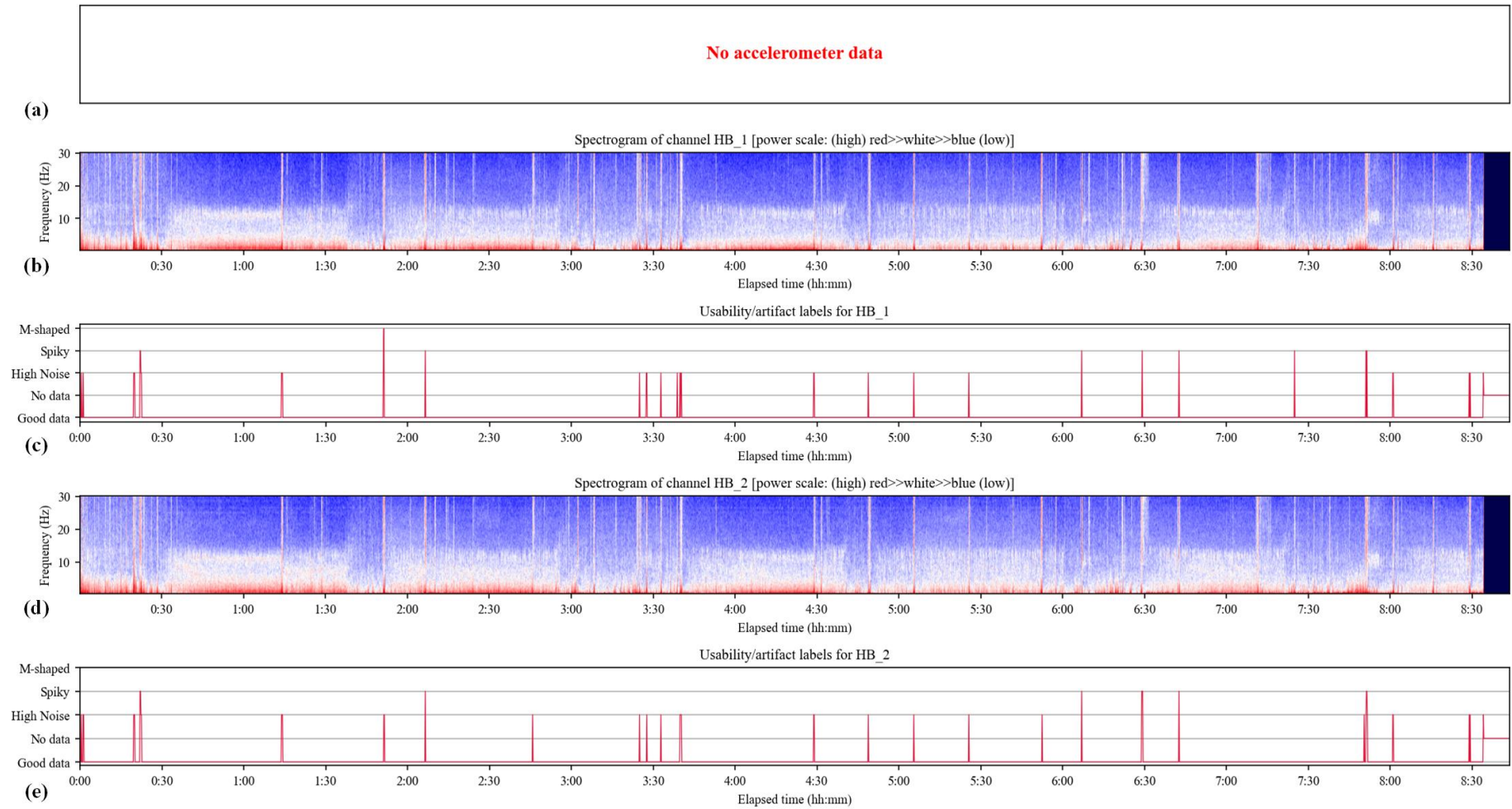

**Figure 13:** Usability graph of a Bitbrain Ikon (EEG headband) recording showing **(a)** normalized ACC outputs (unused in this case), **(b)** windowed spectrograms of the HB_1 channel, **(c)** usability scores for HB_1, **(d)** windowed spectrograms of the HB_2 channel, and **(e)** usability scores for HB_2.

However, since we do not have manual artifact labels for these recordings, performing a direct quantitative comparison of the usability scores is challenging. Nevertheless, based on visual inspection of the usability graphs, the model's outcomes on non-Zmax data appear consistent with those from the Zmax data. In the Supplementary Materials, we have provided over 400 usability graphs of EEG data from the mentioned devices, offering a broader view of the model's performance on non-Zmax datasets. Still, it is always a good idea to review a subset of usability graphs during post-processing to ensure the model's outputs align with the corresponding spectrograms.

## 5. Discussion and limitations

In the previous sections, we described the fundamentals of the eegFloss package and its core models. In this section, we discuss some aspects of the package and the models that are vital for clarity and transparency.

### 5.1. eegUsability: manual labeling considerations

As detailed in Section 3.1.1, we manually labeled EEG artifacts in 181 nights of Zmax data by visually inspecting spectrograms and time-domain signals to prepare the training data of the eegUsability model. However, this procedure was performed by a single scorer (N.S.), which has the risk of introducing subjective biases. Additionally, as the manual labeling progressed, some artifacts were discovered (and deemed significant or frequent enough) after several nights had been scored. However, when a new artifact was defined, previously scored nights were revisited to identify their presence in them. Moreover, since data containing compound artifacts do not always adhere to the clear-cut definitions of the primary artifacts, their identification often depends on the scorer's perception. Therefore, given the extensive volume of data and the intricate nature of the artifacts, the likelihood of labeling inaccuracies (or inconsistencies) cannot be entirely dismissed. Furthermore, it is challenging to assess if our approach of merging samples containing different (compound) artifacts (described in Section 3.1.1.3) has impacted the model's performance. However, we have also introduced a binary classification model in Section 5.3 that only separates good epochs from contaminated ones and does not distinguish between the types of artifacts. The performance of this model is comparable to that of the multi-class classification model, suggesting that the effect of merging samples may have been negligible. Additionally, two scorers are currently working on manually artifact-labeling more EEG recordings, which will be used to train future versions of the eegUsability model, thereby reducing subjective biases in the training dataset.

### 5.2. eegUsability: different epoch lengths

As mentioned in Section 3.1.1.4, the eegUsability model is trained on EEG data segmented in 10s-epochs, and all examples presented thus far have utilized the same window size for usability testing. However, the eegFloss package allows for the assessment of EEG data using different window sizes by adjusting a specific parameter. Figure 14 demonstrates the results of evaluating a night with noisy EEG data using different epoch durations. The figure shows that employing larger window sizes still produces outputs consistent with the nature of the data.

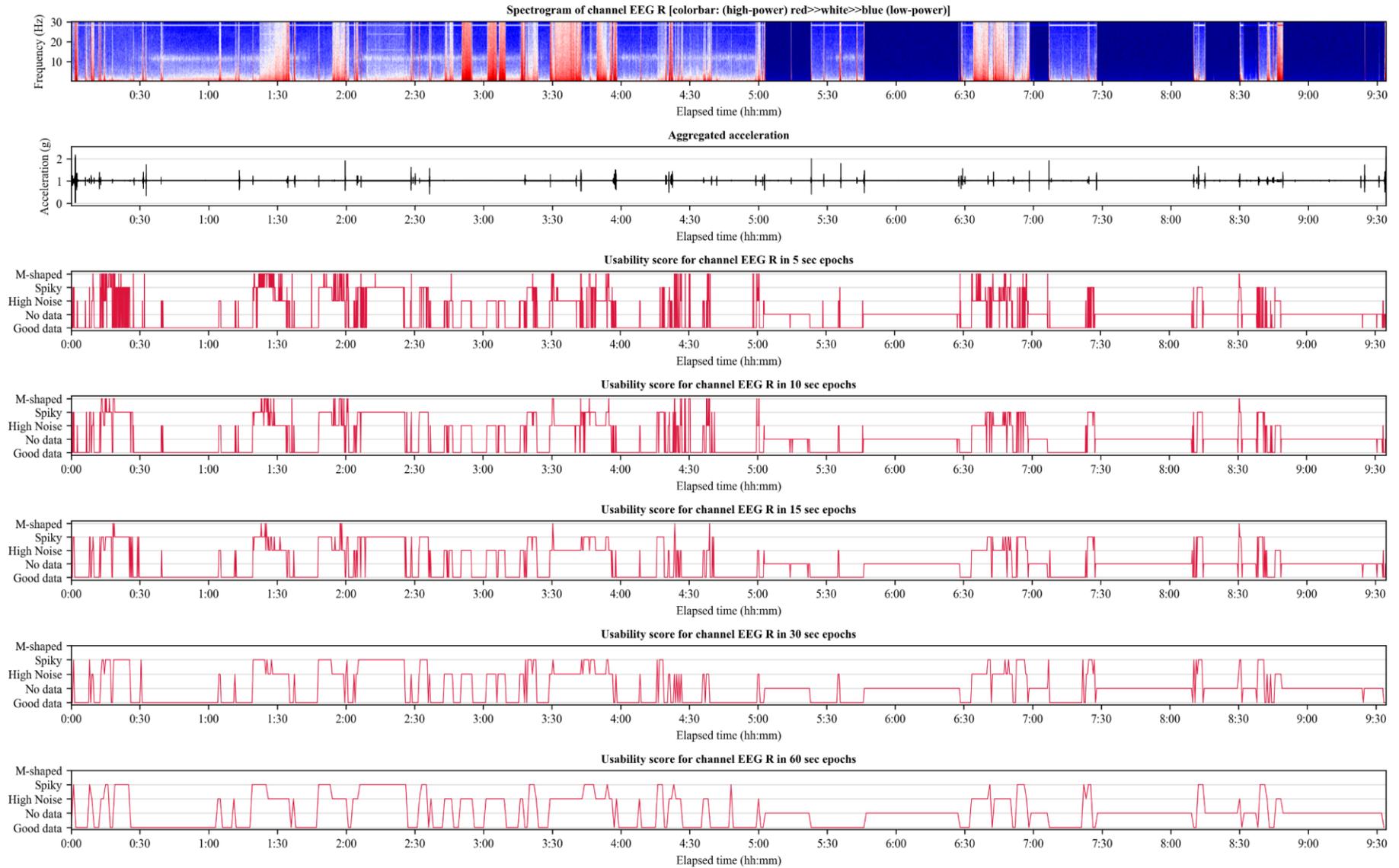

**Figure 14:** **(a)** A sample Zmax EEG channel's spectrogram and its usability scores checked in **(b)** 5, **(c)** 10, **(d)** 15, **(e)** 30, and **(f)** 60-second epochs.

## 5.3. eegUsability: variants and processing time

Checking EEG data using the eegUsability model can be time-consuming. The primary contributor to this factor is the extraction of TSFEL features using the library of the same name (discussed in Section 3.1.3). To address this issue, the model's poor performance in detecting M-shaped Noise (discussed in Section 3.1.5), and the concerns regarding merging different types of artifacts (discussed in Section 5.1), we have developed several variants of the eegUsability model to optimize eegFloss's performance and processing speed. For clarity, we refer to these variants as different versions with specific (names and) version numbers. The primary (or default) eegUsability model (described in Section 3.1) is referred to as "v1.0", and the rest are marked as previous versions. The primary model variants are "weighted-M," which enhances the recall for detecting the M-shaped Noise, and "binary," which simplifies classification into usable and unusable categories, omitting differentiation between artifact types. Additionally, "lite" versions of these primary models utilize only spectrogram features to bypass the time-consuming TSFEL feature extraction. Finally, the "full" version is trained on the entire 127 nights of labeled data, which is 16.5% more than that of the default model (in terms of the number of samples). The rest of the steps associated with model training remained the same as those of eegUsability v1.0. However, the LightGBM parameters of each model were separately hypertuned to obtain optimal performances. Table 4 presents the differences in properties and the training process of these models, along with guidelines on their appropriate usage contexts.

Table 5 outlines each variant's performance metrics and processing time, tested on a system with an Intel Core i7-11850H processor (64-bit, 8C/16T, 2.5–4.8 GHz), 16 GB of RAM, and no GPU. Processing an eight-hour Zmax night with the default model takes ≈35 seconds, including TSFEL feature extraction, classification, and the aggregation of sleep and usability scores. By contrast, the lite models, which work without TSFEL features, demonstrate significantly reduced execution times (a few seconds) compared to the primary models. However, they show a slight decrease in classification performance. Detailed class-wise performance metrics for these models are available in Appendix 3. Although these models provide slightly lower recall scores in identifying usable data than their counterparts, their overall performance remains largely comparable. However, some inconsistencies between the real EEG signals and their usability scores were observed while manually inspecting usability graphs provided by the lite models. Therefore, if processing time is not an issue, using the primary variants that take both sets of features into account is recommended. Overall, eegUsability v1.0 remains the standard for its balanced and reliable performance across various datasets, particularly in accurately identifying usable data. Since the eegMobility model also uses TSFEL features, its processing time is similar to that of eegUsability v1.0. Therefore, we have developed a lite version of the eegMobility model as well, based on Welch's power features. The performance of this model has been presented in Appendix 2.

The processing times can be significantly reduced by increasing the number of available processor cores as well. Regarding space requirements, eegFloss is designed to use working memory efficiently by actively recycling large variables and periodically cleaning residual data. Processing an eight-hour Zmax night (with two EEG channels) typically requires less than 4 GB of RAM. However, processing longer Zmax recordings or data from other devices with more channels will increase RAM usage.

**Table 4:** The variants of the eegUsability model and when to use them.

| eegUsability version | Feature set(s) | Specialty | When to use |
|---|---|---|---|
| "v1.0" or "default" | Spectrogram and statistical | Combines both feature sets for consistent outputs. Tested across datasets and most dependable (described in Section 3.1). | Best for general tasks requiring maximum data retention. |
| "v0.8" or "weighted-M" | Spectrogram and statistical | Better at identifying M-shaped Noise but sacrifices a bit more usable data. | Ideal when M-shaped Noise detection is crucial and slight data loss is acceptable. |
| "v0.6" or "binary" | Spectrogram and statistical | Only identifies whether the data is usable or not; does not differentiate noise types. | If noise differentiation is entirely unnecessary or processing simplicity is prioritized. |
| "v0.7" or "lite" | Spectrogram | Uses only one feature set; similar to v1.0, but 12 times faster with comparable results. | Suitable for quick results where minor inconsistencies are tolerable. |
| "v0.7.2" or "lite weighted-M" | Spectrogram | Similar to v0.8, but works on only spectrogram features, hence is faster. | Optimal for quick, precise outputs. |
| "v0.7.3" or "lite binary" | Spectrogram | Similar to v0.6 but works on only spectrogram features, hence is faster. | Handy when fast results are needed without noise type differentiation. |
| "v0.9" or "full" | Spectrogram and statistical | Similar to v1.0 but is trained on the entire available dataset. | Can be used if a more hypertuned model is needed. |

**Table 5:** Performance of the eegUsability model variants.

| eegUsability version | Weighted mean | | | κ | Training data (%) | Processing time for an 8-hr night^ |
|---|---|---|---|---|---|---|
| | Precision (%) | Recall (%) | F1-score (%) | | | |
| v1.0 (default) | 84.98 | 86.01 | 84.87 | 0.780 | 85.82 | ≈35 sec |
| v0.8 (weighted-M) | 86.33 | 86.45 | 86.35 | 0.791 | 85.82 | ≈36 sec |
| v0.6 (binary) | 89.44 | 89.41 | 89.4 | 0.788 | 85.82 | ≈35 sec |
| v0.7 (lite) | 85.15 | 85.96 | 84.94 | 0.781 | 85.82 | ≈3 sec |
| v0.7.2 (lite weighted-M) | 86.58 | 86.21 | 86.37 | 0.790 | 85.82 | ≈3 sec |
| v0.7.3 (lite binary) | 89.29 | 89.29 | 89.29 | 0.786 | 85.82 | ≈3 sec |
| v0.9 (full) | 90.14* | 90.34* | 90.11* | 0.850* | 100 | ≈37 sec |

*Results are from a test set that is a subset of the training data.
^Tested on a Core i7, 8C/16T, 2.5–4.8 GHz processor with no resource-intensive processes running in parallel.

## 5.4. eegUsability: effects on autoscores

One of the principal goals behind developing eegUsability was to improve the quality of autoscores by identifying and removing artifact-contaminated epochs, before or after autoscoring. To quantify this improvement, we compared (PSG-based) manual scores with (Zmax–based) Dreamento autoscores on two independent datasets: Wearanize+ (100 nights) and QSci (57 nights) [105,106]. Using eegFloss, we determined that 18.95% of Wearanize+ and 4.73% of QSci Zmax EEG epochs contained artifacts. Figure 15 presents detailed stage-wise confusion matrices for both datasets, with and without epochs with artifacts. Prior to artifact rejection, Dreamento achieved overall F1-scores of 48.79% (Wearanize+) and 69.93% (QSci). After excluding epochs flagged by eegFloss, these scores increased to 53.52% and 71.96%, respectively, corresponding to absolute gains of 4.73% and 2.03% in agreement. These results demonstrate that exclusion based on data usability can improve agreements between automated and manual sleep staging, and in turn, the overall quality and reliability of wearable-based autoscores.

## 5.5. eegMobility: validation and shortcomings

As outlined in Section 3.2.3, eegMobility demonstrates nearly perfect classification performance in detecting participants' degrees of mobility. Similarly, the TIB detection algorithm of eegFloss (detailed in Section 4.3) generally performs as expected. However, due to the absence of a quantitative means to assess the success of TIB detection, we manually reviewed its outcomes within the Wearanize+ dataset. Of the 122 available nights, the detected TIB durations were largely accurate for 119 nights, based on the two-minute threshold implemented by eegFloss. However, this simple approach may encounter difficulties under certain conditions, such as prolonged Idle periods within detected Lying phases or if the device stops recording data without terminating the recording. Two examples of such scenarios have been presented in Appendix 4. Nevertheless, if the raw ACC data is consistent, TIB detection by eegFloss is usually quite accurate. The two-minute threshold was chosen based on observed outcomes across various settings, striking an optimal balance between minimizing non-sleep-related data and preserving relevant information. The threshold can be adjusted to suit specific datasets. The effectiveness of eegMobility for TIB detection on non-Zmax data has not yet been extensively tested. Generalization is particularly challenging due to the lack of standardized guidelines regarding ACC sensor placement for sleep monitoring and the data export format. That said, Zmax is typically placed on the forehead, and it stores raw ACC data in g units. If the device follows similar characteristics, eegMobility can be adapted accordingly.

## 5.6. eegFloss: package stability

eegFloss is designed to process recordings from multiple subjects and nights sequentially. Real-world data often presents unforeseen challenges, and eegFloss is well-equipped to handle them. From extensive testing of the package on multiple datasets, we have identified thirteen potential errors (or mismatches) that might occur while processing a night's data. To save time and avoid interruptions, eegFloss is configured to skip the night if any of these errors manifest, proceeding directly to the next recording. A list detailing the unprocessed nights and the exact errors encountered is provided at the end of the script's execution. The eegFloss package has been rigorously tested on Windows (10 and 11), Linux (Ubuntu 24.04), and macOS (Sequoia 15.3.1) and has been applied to more than five Zmax and two non-Zmax datasets.

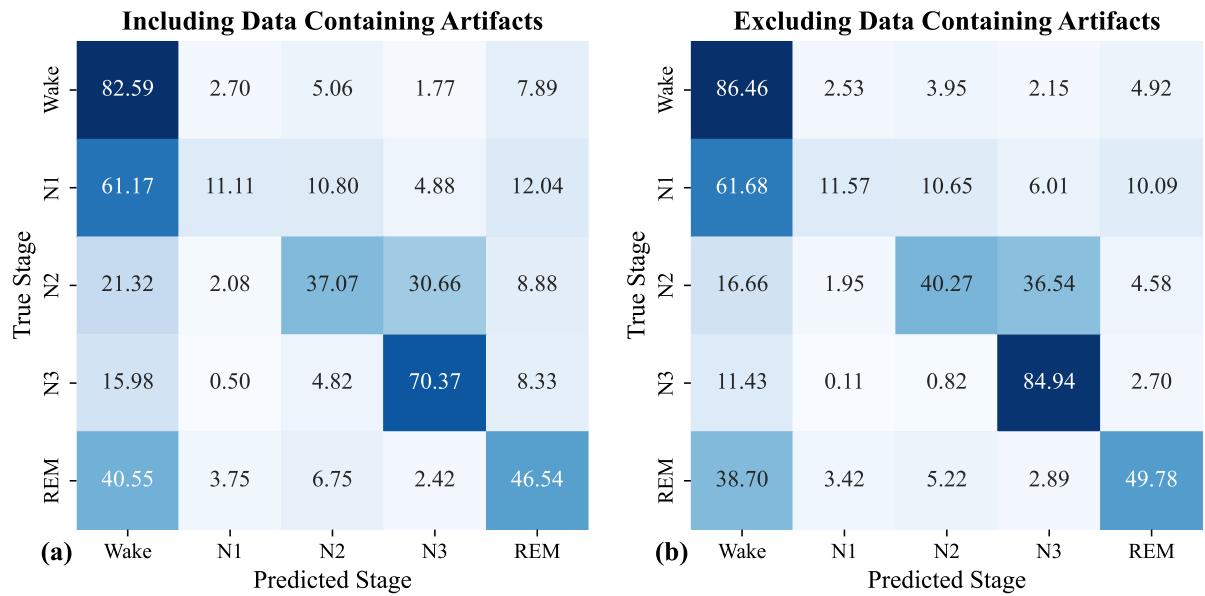

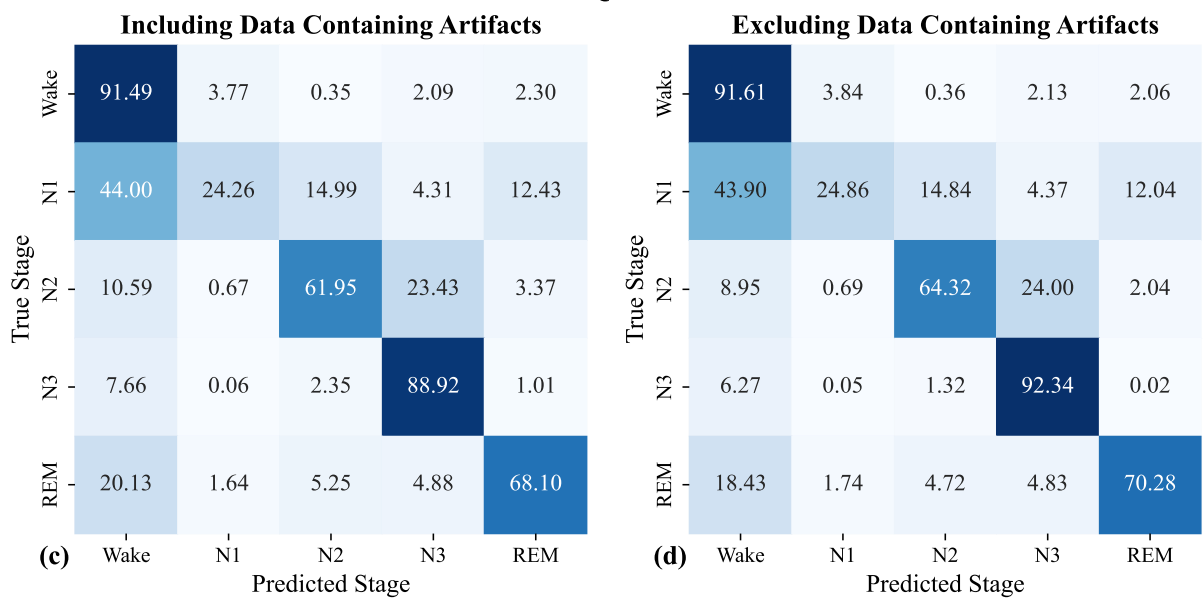

**Figure 15:** Confusion matrices showing the stage-wise agreement between PSG-based manual scores and Zmax-based Dreamento autoscores **(a, c)** without and **(b, d)** with eegUsability-based artifact rejection in Wearanize+ and QSci datasets, respectively.

# 6. Conclusions and future plans

In this article, we introduced eegFloss—an open-source Python package designed to refine sleep EEG recordings by "flossing out" various artifacts and non-sleep-related parts. At the core of eegFloss are two (sets of) novel ML models—eegUsability and eegMobility—both LightGBM-based classifiers. The eegUsability model is trained (and evaluated) on 127 nights of manually artifact-labeled Zmax EEG data from 15 participants. It can detect four types of artifacts (No Data, High Noise, Spiky Noise, and M-shaped Noise) in EEG data from various devices and determines whether data segments are suitable for sleep autoscoring. Notably, it achieves a high recall rate (93.55%) for identifying usable data while maintaining solid overall classification performance ($\kappa = 0.78$), which aids in minimizing the exclusion of good data—a crucial factor for most sleep studies. eegMobility is trained on ≈23 hours of activity data and can identify the participants' degree of mobility throughout the night from Zmax recordings. The resultant mobility scores enable eegFloss to detect TIB precisely. With near-perfect classification performance ($\kappa = 0.994$), this model addresses the need for automatic TIB estimation in sleep research. Both models are available in several variants, each offering distinct advantages tailored to the specific needs and objectives of the analysis.

Beyond artifact detection and TIB identification, the current version of eegFloss can aggregate provided autoscores (or manual scores) with usability scores to obtain artifact-rejected scores, filter out Spiky Noise from affected segments to increase usable data, generate hypnograms, and calculate sleep statistics. The package is fairly easy to set up and use (by following detailed instructions on the code repository page) and can be adapted by modifying specific parameters to meet the needs of different projects. It has been extensively tested on multiple datasets and across the three primary operating systems, demonstrating its robustness and practical utility. From a sleep researcher's perspective, eegFloss presents a novel, straightforward, and elegant solution to the pervasive challenge of dealing with artifacts in EEG data, potentially enhancing both the precision of study analyses and the reliability of their outcomes. Moreover, this tool holds promise for citizen scientists who use consumer-grade headbands to record their sleep at home and often encounter poor data quality or erroneous sleep-stage classifications [43].

In future versions of the eegFloss package, we aim to expand the training dataset for the eegUsability model to enhance its accuracy and generalizability and improve the detection of M-shaped noise. Additionally, we plan to revise the workflow of the eegFloss package (presented in Figure 5) to perform usability detection prior to autoscoring. This adjustment will allow the autoscorer to take the usability scores into account and optimize the autoscoring process accordingly. Alternatively, we are considering developing a new autoscorer capable of identifying the segments with artifacts as a distinct class, thus avoiding the scoring of contaminated data. We also intend to utilize data cleansed of Spiky Noise for autoscoring, thus reducing the loss of usable data. Another significant update will involve optimizing the eegUsability model based on the outcomes of feature extraction—by selecting the most effective TSFEL features (and potentially bypassing the current extraction library), we aim to reduce the processing time, which might also improve model performance. Moreover, we plan to explore other modes of information extracted from EEG signals and incorporate them to improve the eegUsability model. Furthermore, to make eegFloss more comprehensive, future updates will include power calculations across various brainwave bands (such as α, β, γ, and δ), periodogram visualizations, and an extended set of sleep statistics. Lastly, we plan to identify the sources of the artifacts observed in the Zmax EEG data and devise strategies to minimize them at their origin. These efforts will collectively streamline the collection and processing of sleep EEG recordings, leading to even more precise and impactful outcomes from sleep research projects.

## Conflict of interest

The authors declare no conflict of interest.

## Acknowledgments

The authors would like to express their gratitude to all researchers who provided feedback on earlier versions of the eegFloss package.

## Code availability

The eegFloss package is available under the MIT License at [GitHub.com/Niloy333/eegFloss](GitHub.com/Niloy333/eegFloss). For citation, please use DOI: 10.5281/zenodo.15823969.

## Supplementary materials

Generated by eegFloss, 900+ usability graphs and hypnograms from Zmax data and 400+ usability graphs from non-Zmax EEG data across multiple datasets are available here: [eegFloss_outputs](eegFloss_outputs).

# Appendix 1.    Additional noises detected in Zmax data

This section presents sample artifacts of particular classes (described in Section 3.1.1.2) in Figure 1.1. Please take the amplitude of different plots into account while assessing them.

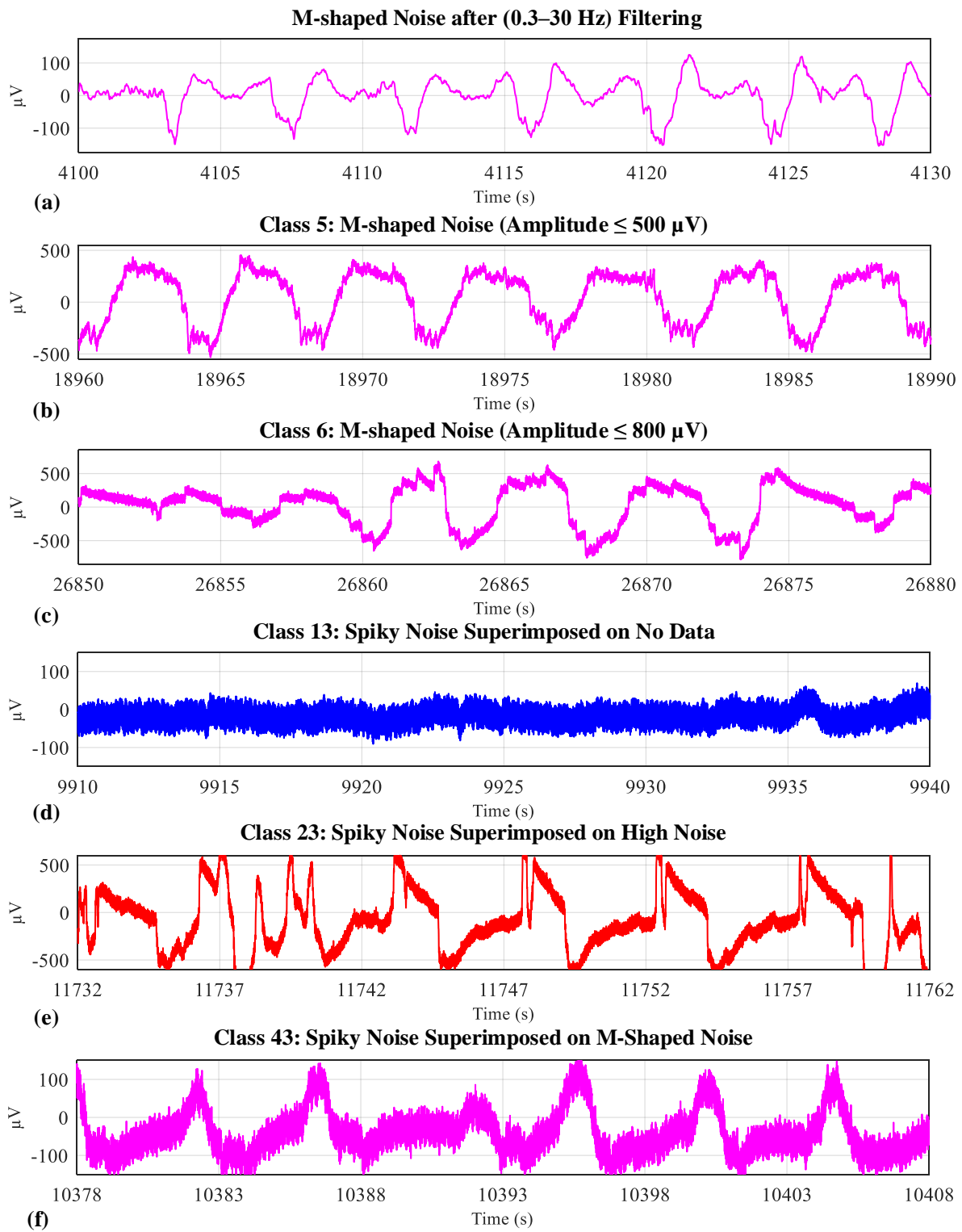

**Figure 1.1: (a)** A sample Spiky Noise after 0.3–30 Hz filtering; M-shaped Noise with amplitudes ranging from **(b)** −400 µV to 400 µV and **(c)** −800 µV to 800 µV, respectively; and Spiky Noise superimposed on **(d)** No Data, **(e)** High Noise, and **(f)** M-shaped Noise. Please note the changes in axes limits in different subplots.

# Appendix 2.  eegMobility: model variations

In this section, we present the performance of the eegMobility lite model that works on power features extracted from tri-axial ACC signals instead of TSFEL features, making it considerably faster. Table 2.1 presents a comparative view of the performance of the default and lite eegMobility models, while detailed class-wise scores for the eegMobility lite model are shown in Figure 2.1. The training process of this model follows the methodology outlined for eegMobility in Section 3.2.

Table 2.1: Performance of different versions of the eegUsability model.

| Model name | Weighted mean | | | κ | Class-wise sample ratio* (%) | Processing time for an 8-hr night^ |
|---|---|---|---|---|---|---|
| | Precision (%) | Recall (%) | F1-score (%) | | | |
| eegMobility (Section 3.2) | 99.64 | 99.63 | 99.63 | 0.994 | 4.60, 57.27, 16.43, 21.70 | ≈32 sec |
| eegMobility lite | 97.98 | 97.98 | 97.98 | 0.966 | 4.60, 57.27, 16.43, 21.70 | ≈2 sec |

*Classes: Idle, Lying, Stationary, and Mobile (both training and test sets).
^Tested on a Core i7, 8C/16T, 2.5–4.8 GHz processor with no resource-intensive processes running in parallel.

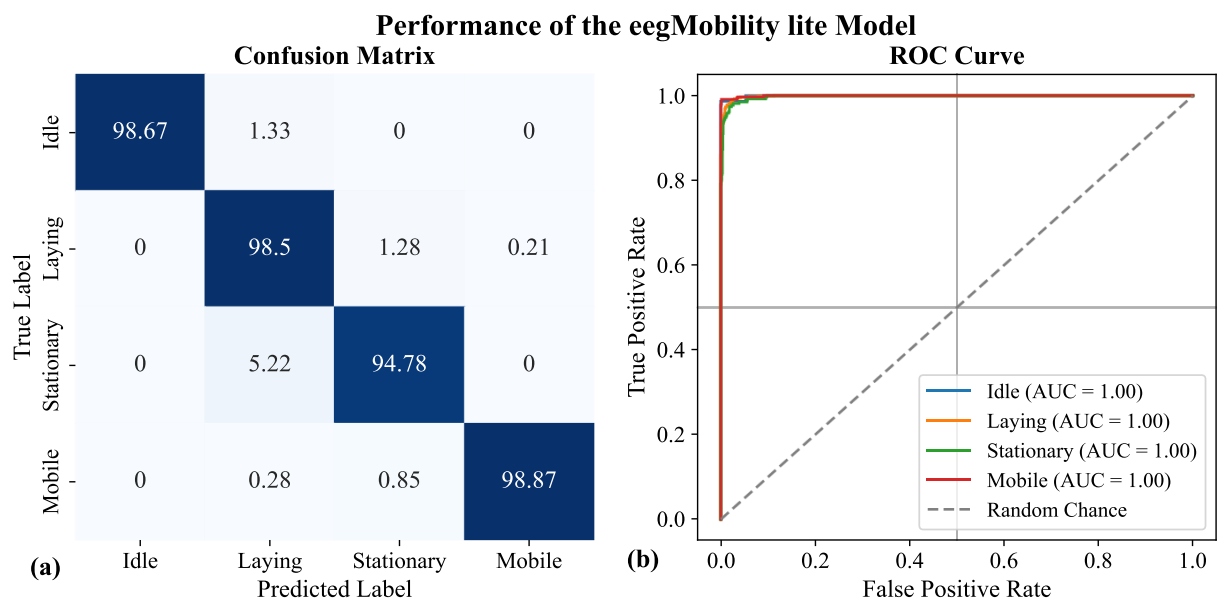

Figure 2.1: Class-wise performance of the eegMobility lite model, expressed in terms of (a) a confusion matrix (in percentages) and (b) ROC curves.

# Appendix 3. eegUsability: model variations

In this section, we detail the performance of the various eegUsability model variants, as previously discussed in Section 5.3. Figures 3.1–3.6 display the confusion matrices and class-wise ROC curves for the eegUsability model versions: v0.8 (weighted-M), v0.6 (binary), v0.7 (lite), v0.7.2 (lite weighted-M), v0.7.3 (lite binary), and v0.9 (full), respectively. The specific attributes and recommended applications of these models have been outlined in Table 4.

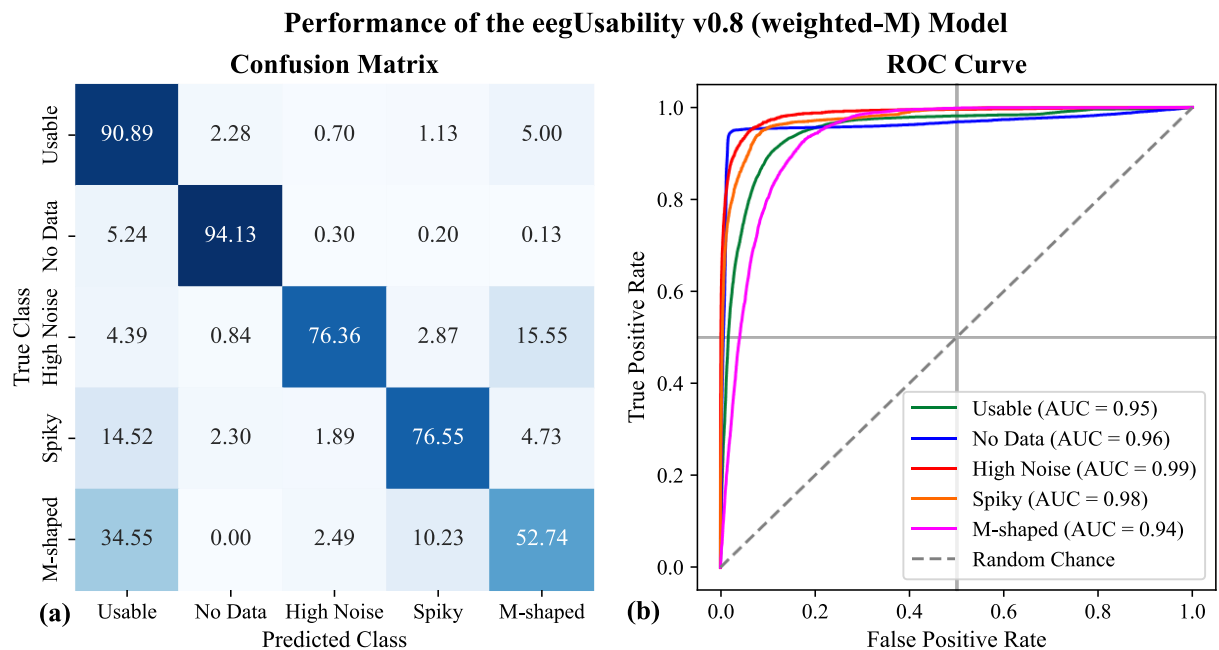

**Figure 3.1:** Class-wise performance of the eegUsability v0.8 (weighted-M) model, expressed in terms of **(a)** a confusion matrix (in percentages) and **(b)** ROC curves.

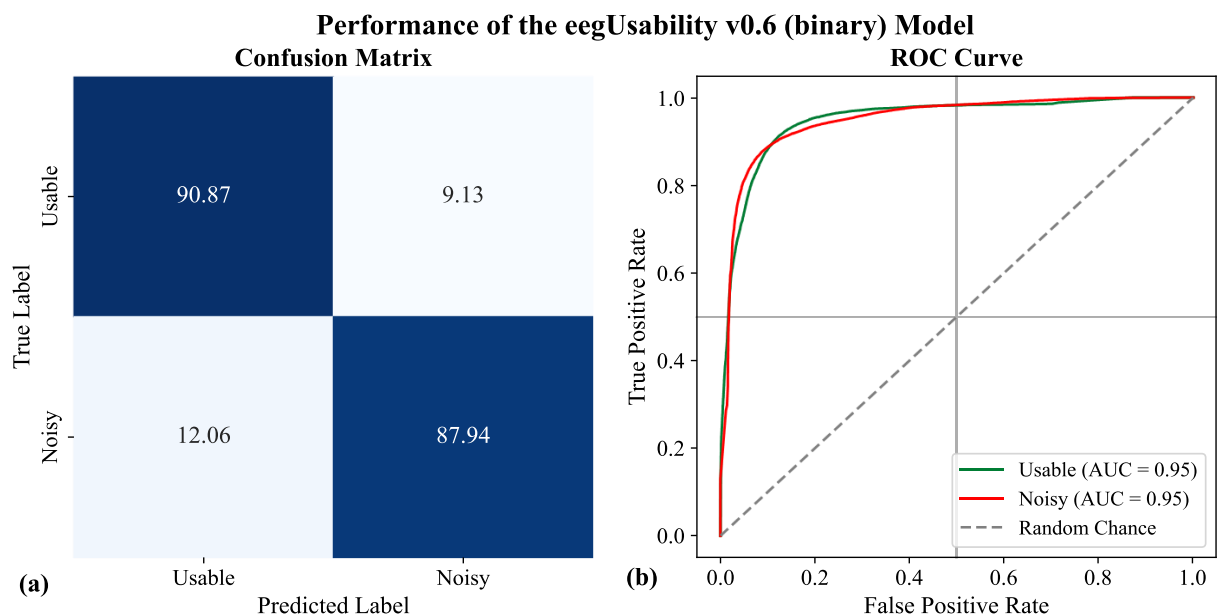

**Figure 3.2:** Class-wise performance of the eegUsability v0.6 (binary) model, expressed in terms of **(a)** a confusion matrix (in percentages) and **(b)** ROC curves.

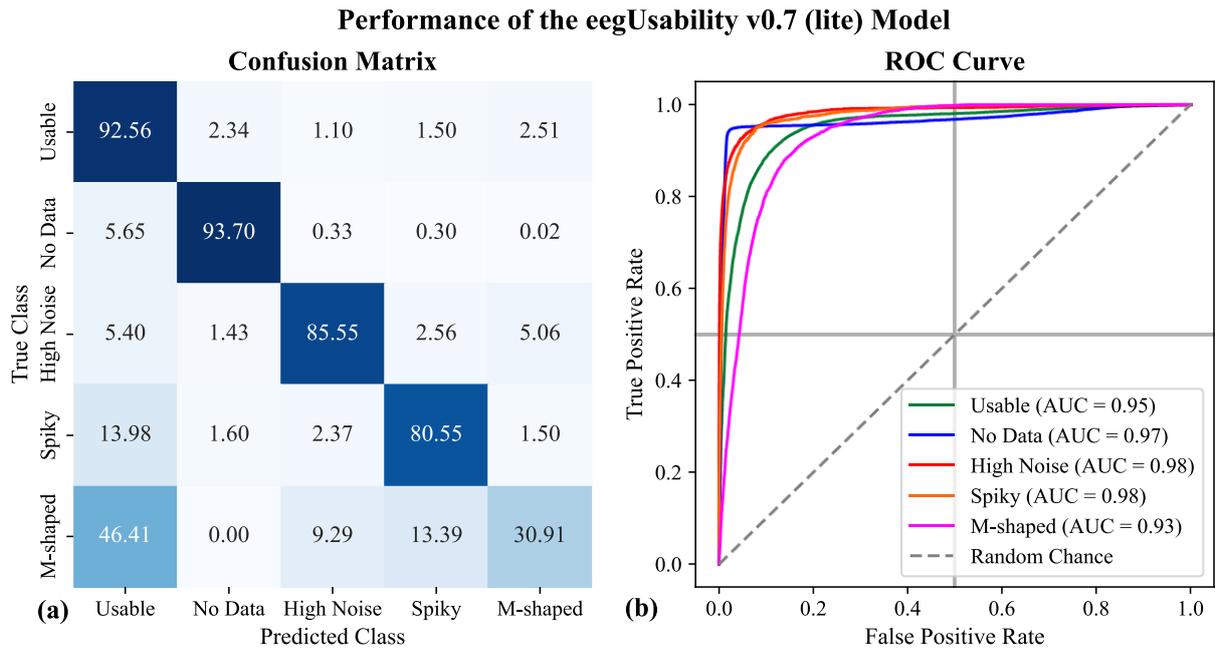

**Figure 3.3:** Class-wise performance of the eegUsability v0.7 (lite) model, expressed in terms of **(a)** a confusion matrix (in percentages) and **(b)** ROC curves.

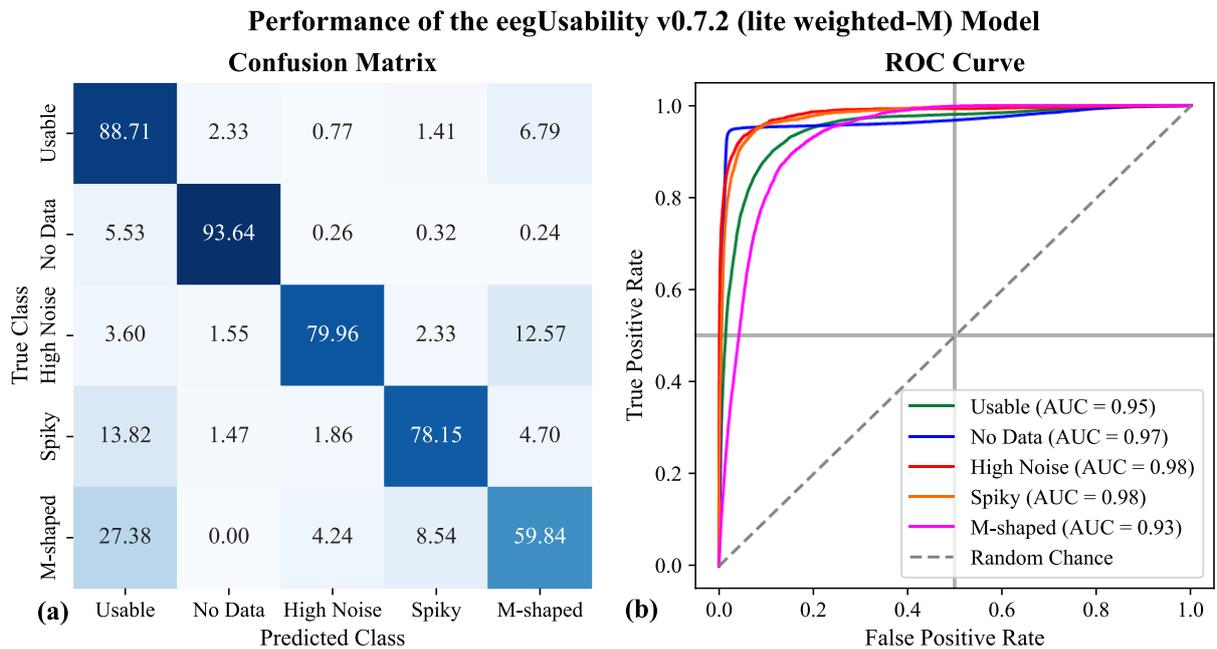

**Figure 3.4:** Class-wise performance of the eegUsability v0.7.2 (lite weighted-M) model, expressed in terms of **(a)** a confusion matrix (in percentages) and **(b)** ROC curves.

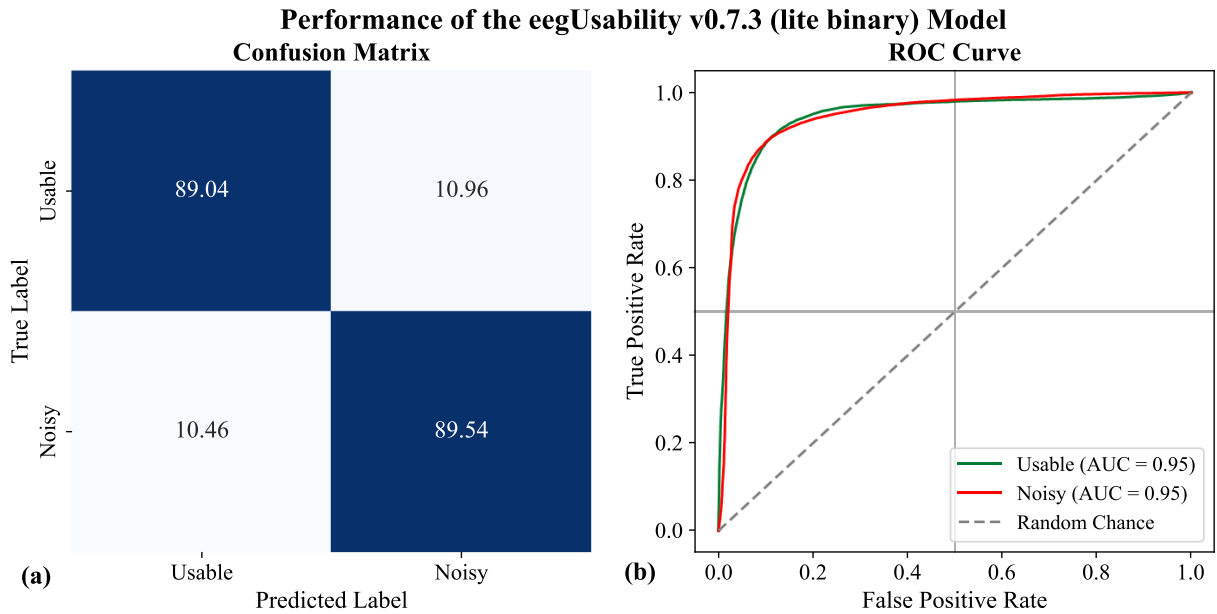

**Figure 3.5:** Class-wise performance of the eegUsability v0.7.3 (lite binary) model, expressed in terms of **(a)** a confusion matrix (in percentages) and **(b)** ROC curves.

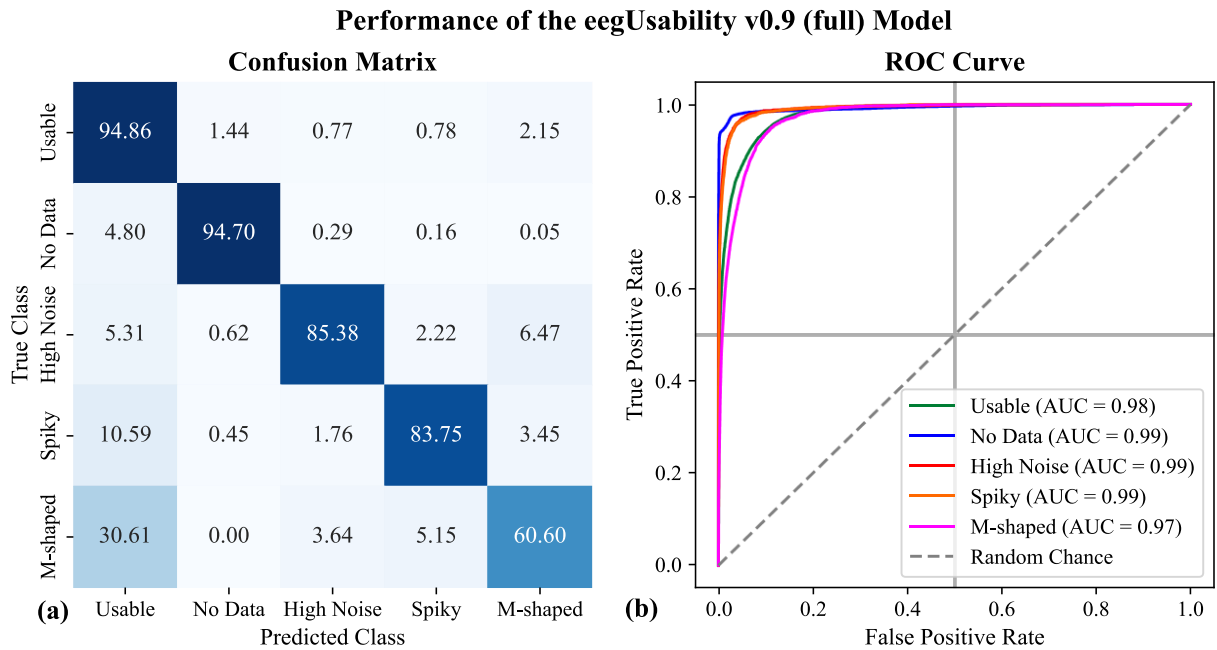

**Figure 3.6:** Class-wise performance of the eegUsability v0.9 (full) model on a subset of the training samples, expressed in terms of **(a)** a confusion matrix (in percentages) and **(b)** ROC curves.

## Appendix 4. eegMobility: failed cases

In this section, in Figures 4.1 and 4.2, we illustrate the instances where eegFloss incorrectly identified TIB based eegMobility outputs due to the nature of the recorded movements.

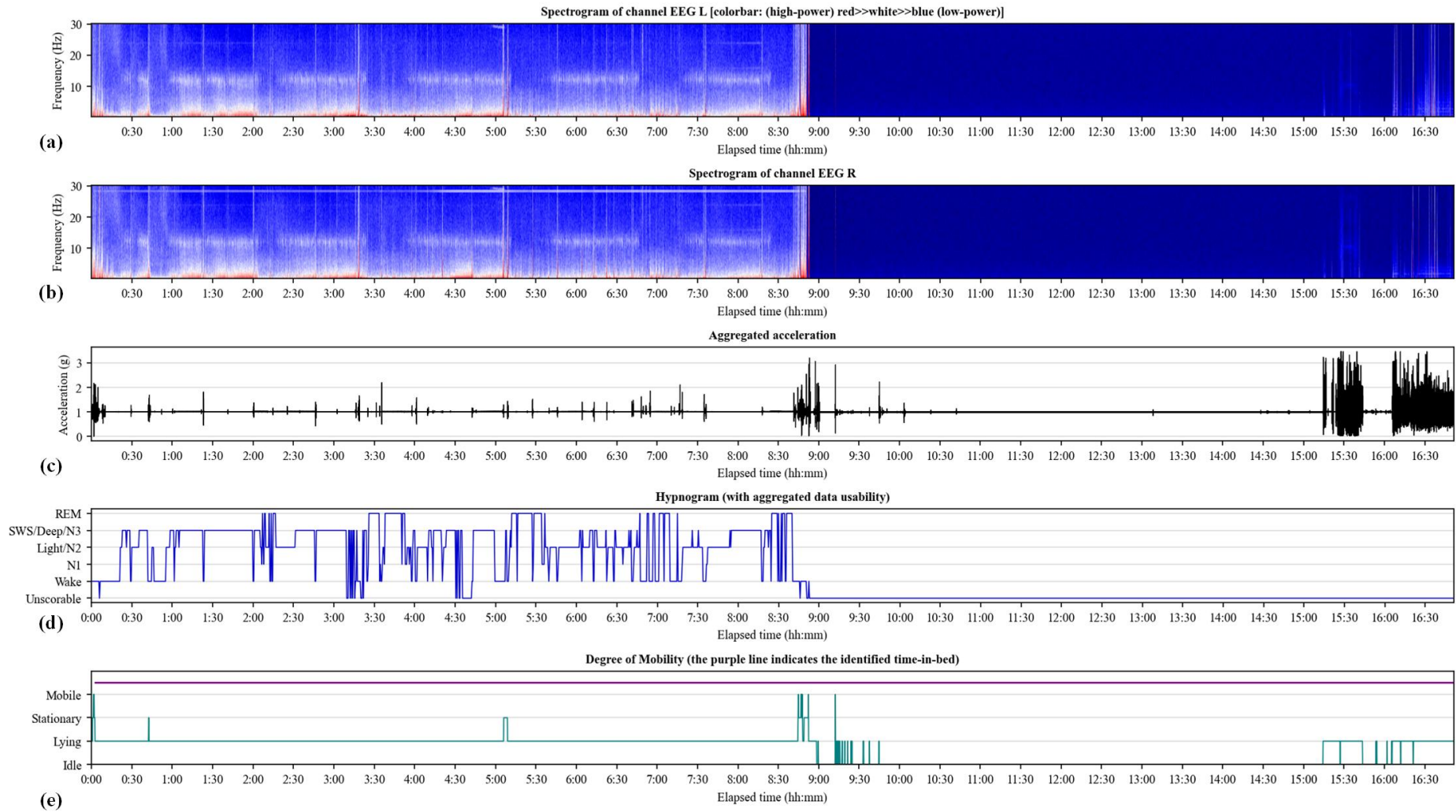

**Figure 4.1:** eegFloss outputs of a sample Zmax recording showing spectrograms of **(a)** EEG Left and **(b)** EEG Right channels, **(c)** the normalized acceleration, **(d)** hypnogram based on the artifact-rejected autoscores, and **(e)** the mobility labels with the detected TIB (which is inaccurate).

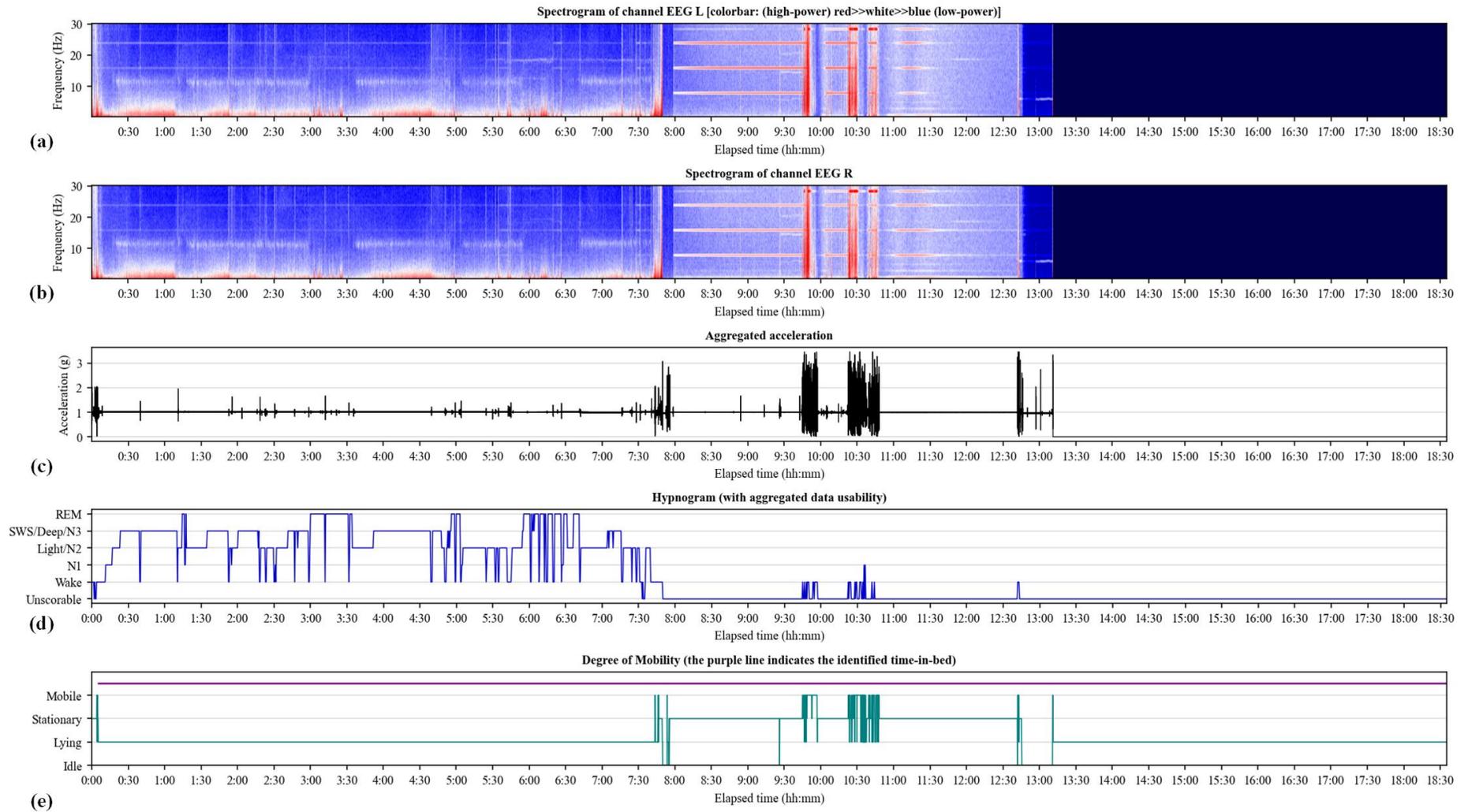

**Figure 4.2:** eegFloss outputs of a sample Zmax recording showing spectrograms of **(a)** EEG Left and **(b)** EEG Right channels, **(c)** the normalized acceleration, **(d)** hypnogram based on the artifact-rejected autoscores, and **(e)** the mobility labels with the detected TIB (which is inaccurate).